\documentclass[letterpaper]{article} 
\usepackage{aaai2026}  
\usepackage{times}  
\usepackage{helvet}  
\usepackage{courier}  
\usepackage[hyphens]{url}  
\usepackage{graphicx} 
\usepackage{natbib}  
\usepackage{caption} 
\usepackage{algorithm}
\usepackage{algorithmic}

\usepackage{newfloat}
\usepackage{listings}
\DeclareCaptionStyle{ruled}{labelfont=normalfont,labelsep=colon,strut=off} 
\floatstyle{ruled}
\newfloat{listing}{tb}{lst}{}
\floatname{listing}{Listing}
\usepackage{amsmath}  
\definecolor{green}{RGB}{0,128,0}  
\definecolor{red}{RGB}{255,0,0}    
\usepackage{multirow}
\usepackage{amsmath}
\usepackage{amssymb}
\usepackage{array}
\usepackage{longtable}
\usepackage{tabularx}
\usepackage{supertabular}
\usepackage{makecell}
\usepackage[capitalize,noabbrev]{cleveref}
\usepackage{enumitem}
\usepackage{subcaption}
\usepackage{pifont}
\usepackage{dsfont}
\usepackage[table,xcdraw]{xcolor}
\usepackage{xspace}
\usepackage{bm}
\usepackage{booktabs}
\usepackage{microtype}

\usepackage{mathtools}
\usepackage{graphicx}

\usepackage{caption}

\newcommand{\mone}{MedS$^3$\xspace}
\newcommand{\gm}{V_\theta}
\newcommand{\pom}{\pi}

\definecolor{myred}{HTML}{C00000}  
\definecolor{mygreen}{HTML}{548235}  
\usepackage[most]{tcolorbox}
\newtcolorbox{promptbox}[2][]{
    enhanced,
    colback=gray!5,
    colframe=gray!50,
    boxrule=0.5pt,
    arc=3pt,
    fontupper=\ttfamily,
    title={#2},
    listing only,
    listing options={
        basicstyle=\small\ttfamily,
        breaklines=true,
        numbers=left,
        numberstyle=\tiny\color{gray},
    },
    #1
}
\newcommand{\blue}[1]{\textcolor{black}{#1}}
\newcommand{\setblue}{\color{black}}
\title{\mone: Towards Medical Slow Thinking with Self-Evolved Soft Dual-sided Process Supervision}

\author {
    Shuyang Jiang\textsuperscript{\rm 1, \rm 3},
    Yusheng Liao\textsuperscript{\rm 2, \rm 3},
    Zhe Chen\textsuperscript{\rm 2, \rm 3},
    Ya Zhang\textsuperscript{\rm 2, \rm 3, }\thanks{Corresponding Authors},
    Yanfeng Wang\textsuperscript{\rm 2, \rm 3},
    Yu Wang\textsuperscript{\rm 2, \rm 3, }\footnotemark[1]
}
\affiliations {
    \textsuperscript{\rm 1}Fudan University\\
    \textsuperscript{\rm 2}School of Artificial Intelligence, Shanghai Jiao Tong University\\
    \textsuperscript{\rm 3}Shanghai Artificial Intelligence Laboratory\\
    shuyangjiang23@m.fudan.edu.cn, \{liao20160907,chenzhe2018,ya\_zhang, wangyanfeng622, yuwangsjtu\}@sjtu.edu.cn
}

\usepackage{bibentry}

\begin{document}

\maketitle

\begin{abstract}
Medical language models face critical barriers to real-world clinical reasoning applications. 
However, mainstream efforts, which fall short in task coverage, lack fine-grained supervision for intermediate reasoning steps, and rely on proprietary systems, are still far from a versatile, credible and efficient language model for clinical reasoning usage.
To this end, we propose \mone, a self-evolving framework that imparts robust reasoning capabilities to small, deployable models. 
Starting with 8,000 curated instances sampled via a curriculum strategy across five medical domains and 16 datasets, we use a small base policy model to conduct Monte Carlo Tree Search (MCTS) for constructing rule-verifiable reasoning trajectories. 
Self-explored reasoning trajectories ranked by node values are used to bootstrap the policy model via reinforcement fine-tuning and preference learning. 
Moreover, we introduce a soft dual process reward model that incorporates value dynamics: steps that degrade node value are penalized, enabling fine-grained identification of reasoning errors even when the final answer is correct.
Experiments on eleven benchmarks show that \mone outperforms the previous state-of-the-art medical model by +6.45 accuracy points and surpasses 32B-scale general-purpose reasoning models by +8.57 points.
Additional empirical analysis further demonstrates that \mone achieves robust and faithful reasoning behavior.
\end{abstract}

\begin{links}
    \link{Code}{https://github.com/pixas/MedSSS}
\end{links}

\section{Introduction}
\label{intro}
\begin{table*}[tbp]
  \centering
  \setlength{\tabcolsep}{1mm}
  \small
    \begin{tabular}{lccccc}
    \toprule
    \multicolumn{1}{l}{\multirow{2}[2]{*}{\textbf{Models}}}  & \textbf{Without } & \textbf{Diverse} & \multirow{2}[2]{*}{\textbf{Small Size}} & \textbf{Reasoning} & \textbf{Process} \\
           & \textbf{Close-sourced Teacher} & \textbf{Clinical Coverage} &   &  \textbf{Specialized}  & \textbf{ Supervision} \\
    \midrule
    UltraMedical & \textcolor{myred}{\ding{55}} & \textcolor{mygreen}{\ding{51}} & \textcolor{mygreen}{\ding{51}} & \textcolor{myred}{\ding{55}} & \textcolor{myred}{\ding{55}} \\
    HuatuoGPT-o1 &  \textcolor{myred}{\ding{55}} & \textcolor{myred}{\ding{55}} & \textcolor{mygreen}{\ding{51}} & \textcolor{mygreen}{\ding{51}} & \textcolor{myred}{\ding{55}} \\
    O1-journey Part 3  & \textcolor{myred}{\ding{55}} & \textcolor{myred}{\ding{55}} & \textcolor{myred}{\ding{55}} & \textcolor{mygreen}{\ding{51}} & \textcolor{myred}{\ding{55}} \\
    m1-7B-32K & \textcolor{myred}{\ding{55}} & \textcolor{myred}{\ding{55}} & \textcolor{mygreen}{\ding{51}} & \textcolor{mygreen}{\ding{51}} & \textcolor{myred}{\ding{55}} \\
    \midrule
    \mone  & \textcolor{mygreen}{\ding{51}} & \textcolor{mygreen}{\ding{51}} & \textcolor{mygreen}{\ding{51}} & \textcolor{mygreen}{\ding{51}} & \textcolor{mygreen}{\ding{51}} \\
    \bottomrule
    \end{tabular}%
  \caption{Comparison of \mone with other medical models. \mone supports flexible inference-time scaling on resource-constrained devices, as well as process reward-guided decoding algorithms without supervision from large proprietary models. }
  \label{tab:compare_other_medm}%
\end{table*}%

Large Language Models (LLMs) have demonstrated significant potential in the medical domain \citep{singhal2023large, nori2023capabilities}, supporting tasks from clinical note generation~\citep{biswas2024intelligent,jung2024enhancing} to precise diagnosis \citep{tu2024towards,liao2024medcare}. Despite these advances, 
accurate reasoning is steadily fundamental to clinical decision-making, where diagnostic and treatment recommendations must be grounded in coherent, evidence-based logic chains~\citep{cabral2024clinical,tordjman2025comparative}. 
Increasing efforts to enhance reasoning capabilities through chain-of-thought~\citep{wei2022chain}, preference learning~\citep{rafailov2023direct} and reinforcement learning~\citep{guo2025deepseek} highlight that correct answers alone are insufficient without trustworthy reasoning processes. 

While existing approaches have demonstrated notable performance, two challenges persist. 
First, the training data used in many studies mostly consist of multiple-choice problems~\citep{huang2025o1,huang2025m1UnleashPotential}, which lacks sufficient diversity and scale, and hence limits model robustness across different domains. 
Second, a growing dependence on large-scale proprietary models~\citep{huang2025o1} introduces practical and ethical considerations.
Although models distilled from these hyper-scale teachers achieve strong performance, they inherit the unverifiability and potential hallucinations~\citep{xu2024hallucination} of their teachers, offering little control over reasoning faithfulness. 
Moreover, the reliance on distillation from external resources would lead to uncontrollable privacy protection for real-world applications.
These challenges highlight a core problem: how to efficiently induce robust, interpretable, and stepwise supervision reasoning in small-scale medical models without relying on proprietary models or noisy synthetic supervision.

To bridge this gap, we propose \mone, a self-evolving medical reasoning framework that enables small models to iteratively improve through Monte Carlo Tree Search (MCTS)-guided exploration and rule-verified refinement. 
Starting from a diverse, curriculum-sampled dataset spanning five medical domains and 16 medical datasets, \mone generates reasoning trajectories with explicit node value estimates, allowing selection of high-quality paths for policy model bootstrapping via reinforcement fine-tuning and preference learning. 
Crucially, we introduce a soft dual-sided process reward model that labels intermediate steps not only by potential correctness but also by value consistency—penalizing steps that degrade node value and marking them as incorrect if the adjusted value falls below zero. 
This enables faithful supervision even in trajectories with correct final answers. 
Table~\ref{tab:compare_other_medm} highlights these advantages in robust long-chain reasoning and breadth of application.

Extensive experiments on eleven clinical reasoning benchmarks and three out-of-domain datasets demonstrate that \mone achieves state-of-the-art performance, outperforming both comparable-sized medical models and much larger general reasoning models, while maintaining superior interpretability and clinical task coverage. 
In summary, our contributions are:

\begin{enumerate}
\item \textbf{Pioneering Step-Level Framework for Medical AI:} We introduce a self-evolution framework that equips small-scale medical models with robust long-chain reasoning via step-level supervision, tailored for a wide range of clinical applications.
\item \textbf{Novel PRM Training Pipeline:} We propose a unique process reward model trained with soft dual-sided labels, which precisely evaluates each reasoning step by jointly predicting future rewards and assessing atomic step necessity, reflecting clinical reasoning’s incremental confidence building and fewer hallucinations.
\item \textbf{State-of-the-Art Clinical Reasoning Performance:} Our self-evolved system \mone significantly surpasses all equal-parameter competitors and larger reasoning models across multiple clinical benchmarks, driven by fine-grained PRM-guided reasoning enhancement.
\end{enumerate}


\section{\mone} 
\begin{figure*}[t]
    \centering
    \includegraphics[width=0.9\linewidth]{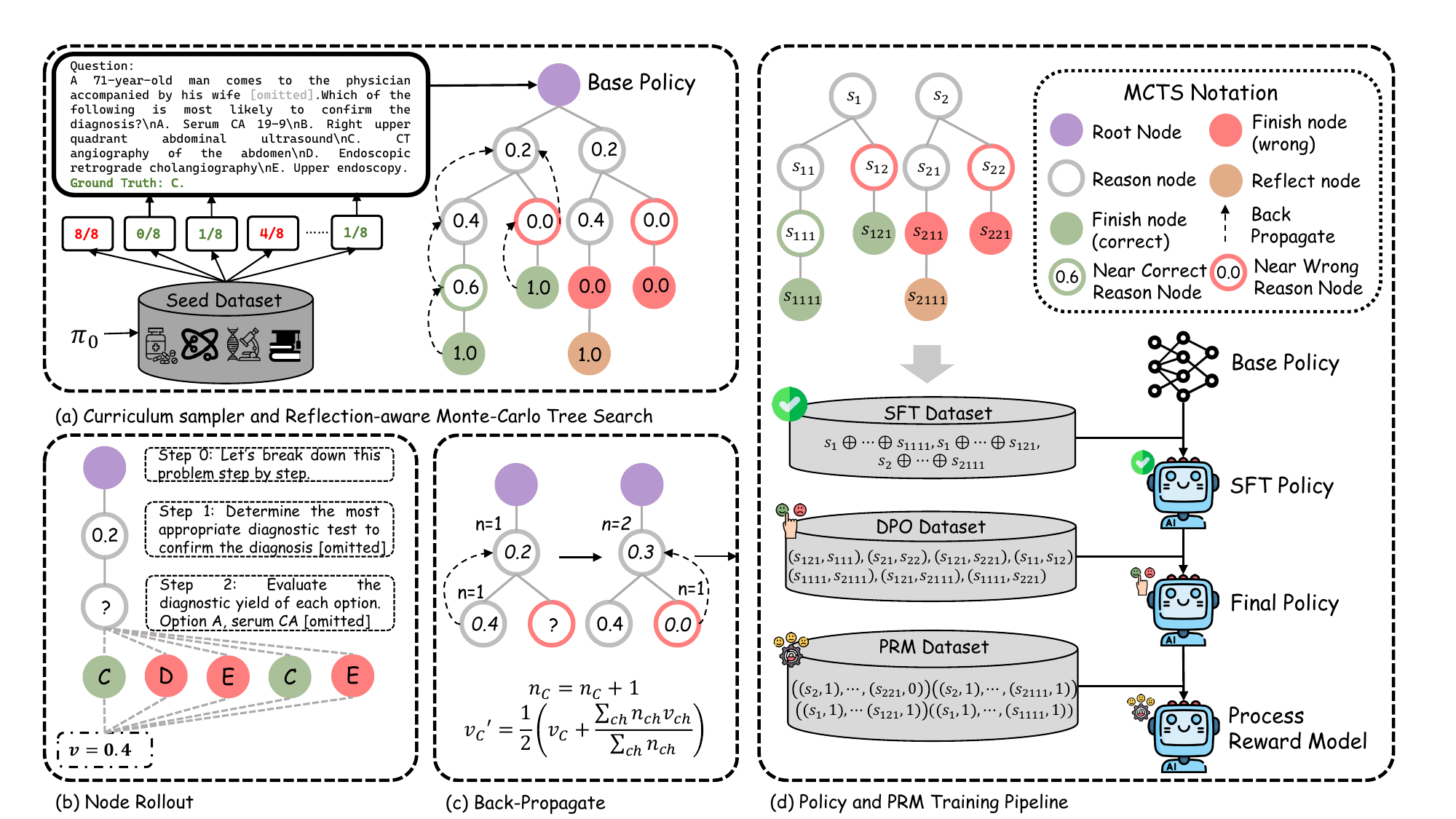}
    \caption{\blue{Overview of the construction of \mone framework. (a) \mone utilizes a Monte-Carlo Tree Search pipeline to self-generate step-by-step reasoning paths for each instance sampled in a curriculum manner. (b) During this process, \mone uses result simulation to obtain the rollout value for each node; (c) After obtaining the child's rollout value, \mone executes back-propagation to enable precise value prediction from deeper layers to transfer back to shallow nodes. (d) After the exploration finishes, we use SFT and DPO to optimize the policy model and soft dual-side label to fine-tune the process reward model.}}
    \label{fig: framework}
\end{figure*}

This section presents a detailed overview of the proposed \mone framework, which is presented in Fig.~\ref{fig: framework}. It is structured into four components:

\begin{enumerate}
    \item \textbf{Self-Bootstrapping Evolution}~(\S\ref{mcts}) which synthesizes reasoning trajectories as training data, with Monte-Carlo Tree Search~(MCTS) technique using the base policy $\pi_0$.
    \item \textbf{Policy Model $\pi$}~(\S\ref{policy_tuning}) which is derived by fine-tuning on the generated synthetic data with supervised learning and direct preference optimization~\citep{rafailov2023direct}.
    \item \textbf{Process Reward Model~(PRM) $V_{\theta}$}~(\S\ref{grader_tuning}) which is fine-tuned with step-wise supervision using soft dual-side labels and assigns a value in the range $[0,1]$ to each reasoning step by a both forward and backward view.
    \item \textbf{Iterative Training Pipeline}~(\S\ref{pipeline}) which consists of two MCTS evolution iterations and a curriculum data sampler.
\end{enumerate}

\subsection{MCTS-guided Evolution}
\label{mcts}
This algorithm builds upon an $n$-ary tree, where 
every root node is initialized as a multi-step reasoning start $s_0=$``Let
s break down this problem step by step.''. 
There are four stages in a full MCTS pipeline, including \textit{Node Selection}, \textit{Node Expansion}, \textit{Node Rollout}, and  \textit{Backpropagation}.

\paragraph{Node Selection}
Within each iteration, we use UCB~\citep{winands2008monte} as the criterion to select a child $T$, which is as follows:
\begin{equation}
    \mathrm{UCB}_T=v_C+\gamma\sqrt{\frac{\ln n_{T_{parent}}}{n_{T}}},
\end{equation}
where $T_{parent}$ is the preceding node of the current node $T$, $n_T$ is the node visiting count, $v_C$ is the node's value obtained by node rollout and updated by back-propagation, and $\gamma$ is an exploration constant set as 2.
For each parent, we select its child node with the highest $\mathrm{UCB}$ value. 

\paragraph{Node Expansion}
After reaching the candidate node $T_c$ under the UCB criterion, we continue the reasoning trace of the current node.
If the current node possesses a relatively high value ($v_c\geq thr$, where $thr=0.9$ is a pre-defined threshold), we prompt the node to directly generate until deriving an answer for speeding up exploration.
For a wrong node, we allow one reflective action \texttt{Reflect} to elicit the introspection of the policy.
Otherwise, assume that the selected node is located at $k$-th layer of the tree with previous reasoning trajectories $[s_{0}, s_1, \cdots, s_k]$ connected by a coherence phrase $t_s$, we sample $B$ subsequent steps $\{s_{k+1,i}\mid i=1,2,\cdots,B\}$ based on the previous trajectory using a \texttt{Reason} node:
\begin{equation}
\label{node_expand}
    s_{k+1,i}\sim \pom_0([s_{0}\oplus s_1\oplus \cdots\oplus s_k]\mid x),
\end{equation}
where $\oplus$ is the operation to connect two steps using the coherence phrase $t_s$, $\pom_0$ is the base policy model, and $x$ is the original input prompt.

\paragraph{Node Rollout}
As the PRM is not yet accurate enough to serve as a reliable critic, node values are obtained using rollouts based on reasoning trajectories so far.
Specifically, for a chosen unvisited node $T_c$ at the $k$-th depth, we set a simulation budget $L=\max(3,\frac{L_0}{k})$ where $L_0=15$, to encourage sufficient simulation trials when the known reasoning path is short, but expect to see a deterministic reasoning result conditioning on a long trajectory.
After setting the budget, we prompt the policy model $\pi_0$ to directly output the answer $L$ times under a specific prompt \verb|AnsPrompt|:
\begin{equation}
\label{rollout}
    a_c^l\sim \pi_0([s_{0}\oplus s_1\oplus \cdots\oplus s_k]\mid x_{\mathrm{AnsPrompt}}),
\end{equation}
where $l\in[1,L]$ and $a_c^l$ is the $l$-th simulated answer.
The average accuracy of the $L$ simulations $acc=\frac{1}{L}\sum_{l=1}^L\mathds{1}_{a_c^l=y}$ is assigned as the value of $T_c$.

\paragraph{Backpropagation}
After the rollout stage, we conduct back-propagation starting from $T_c$ till the root, updating all tree node values along the trace.
Specifically, for an arbitrary node $T_k$, we propose to update its visits $n_k$ and $v_k$ as follows:
\begin{align}
\label{backpropagate}
    n_k&=n_k+1\nonumber\\
    v_k&=\frac{1}{2}\left(v_k+\frac{\sum_{ch}v_{ch}\cdot n_{ch}}{\sum_{ch}n_{ch}}\right),
\end{align}
which considers both correctness and completeness for the evaluation of a reasoning step.

\paragraph{Termination of Search}
To balance the exploration cost and optimization of policy and reward models, we set two criteria to terminate the exploration.
First, once the total correct count in the tree exceeds a minimum correct count $\tau=3$, we stop the exploration of this tree.
Second, if there are no correct nodes after affording a certain number of node exploration trials, we prompt $\pi_0$ to generate \texttt{Finish} node for all leaves.

\subsection{Policy Model Fine-tuning}
\label{policy_tuning}
The policy training first leverages the correct leaves $s_l$ whose values equal 1 and corresponding reasoning trajectories gathered before: $D_{\pom}=\{\left(x,[s_0\oplus s_1\oplus\cdots\oplus s_l]\right)\mid v_l=1\}$.
These correct reasoning traces are fine-tuned to deduce a self-improved policy model:
\begin{equation}
    \label{policy_tuning_formula}
    \mathcal{L}_{\pom}= -\mathbb{E}_{(x,y)\sim D_\pi}\log p_{\theta}(y\mid x),
\end{equation}
where $y=[s_0\oplus s_1\oplus\cdots\oplus s_l]$ is the whole trajectory.
For the second iteration, we further add a step-level Direct Preference Optimization (DPO) to optimize the policy at the same reasoning budget:
\begin{align}
    \mathcal{L}_{\mathrm{DPO}}=-\mathbb{E}_{(x,P^+,P^-)\sim D_{\mathrm{DPO}}}&\log \sigma(r_{\theta}(x,P^+)\nonumber\\
    &-r_{\theta}(x,P^-)),
\end{align}
where $r_{\theta}(x,P)=\beta(\log \pi_{\theta}(P\mid x)-\log \pi_{ref}(P\mid x))$ is the reward and $D_{\mathrm{DPO}}=\{(x,[s_0\oplus s_1\oplus\cdots\oplus s_k^+],[s_0\oplus s_1\oplus\cdots\oplus s_k^-])\mid v_k^+>v_k^-\}$.
DPO training is crucial for deriving a strong policy and PRM, which is elucidated in Table~\ref{tab:ablation}.

\subsection{Soft Dual-side PRM Fine-tuning}
\label{grader_tuning}

\paragraph{Dataset Collection}
We first filter out those trees with only correct or incorrect leaves as these trajectories contain extreme value bias.
For a \texttt{Finish} leaf $T_l$ in a valid tree, its reasoning trace $[(s_1,v_1),(s_2,v_2),\cdots,(s_l,v_l)]$ is one training sample, where each reasoning step is concatenated by ``\texttt{Step k:}'' to form a complete reasoning trajectory.
At the end of each reasoning step $s_i$~(typically a \verb|\n\n| token), the value $v_i$ is used to derive the token label, \blue{which is learned by conditioning on all previous steps in an auto-regressive manner.}
As a result, the PRM training set is such $D_{\gm}=\{(x,[(s_1,v_1),(s_2,v_2),\cdots,(s_l,v_l)])\mid x\in D_{seed}\land s_l\mathrm{\ is\ finish}\}$.

\paragraph{Learning objective} 
Instead of fitting the node value~\citep{zhang2024rest} or learning the pair-wise ranking preference~\citep{guan2025rstar}, we choose to use a binary cross-entropy loss to optimize the PRM for its stability.
Although \citet{zhang2025lessons} suggests that the PRM label should be set to True once the rollout score is above zero, we deem that the rollout score as a soft label has a forward-only bias about reasoning correctness.
A wrong intermediate step is still possible to derive a correct answer given correct prefixes, but such hallucinations are not what medical reasoning desires.
Therefore, a new step is valued highly only when it can both possibly derive a final answer and improve the correctness of the reasoning trajectory deterministically.
As a result, we design a dual-side label $y_i$ for step $i$ using its soft Q-value obtained during MCTS as 
\begin{equation}
\label{eq:soft_dual_side_label}
    y_i=\begin{cases}
        \lceil v_i-\beta \cdot \max(0,v_{i-1}-v_{i+1})\rceil & v_i<v_{i-1} \\
        \lceil v_i \rceil & otherwise
    \end{cases}
\end{equation}
\blue{
This learning objective encourages PRM to simultaneously look ahead and back to judge the current step and penalize intermediate errors except for valid reflection.
}
Based on these, we optimize $\gm$ using the following loss function:
\begin{equation}
    \label{grader_tuning_formula}
    \mathcal{L}_{\gm}=\mathbb{E}_{T_k\sim D_{\gm}}\sum_{i=1}^k y_i \log \hat{y}_i+(1-y_i)\log (1-\hat{y}_i),
\end{equation}
where $\hat{y}_i$ is the predicted probability of the given step $i$ and $\beta$ is a hyperparameter set to 1.0 by a simple grid search (details in Appendix C.1).
This dual-sided soft-label training, not only prevents the learning of fuzzy labels (rollout value around 0.5) but also learns to judge a misleading step.

\subsection{Training Pipeline}
\label{pipeline}
We perform two iterations for the seed dataset.
For each iteration, we use \textbf{curriculum sampler}, which first prompts the policy model to perform the rejected-sampling on the training set, filtering those training instances with all-correct responses to enhance data efficiency.
After that, we sample instances with the lowest average accuracy values during the rejected-sampling process, ensuring that the extremely hard problems (0 accuracy score) are no more than one-third of the total samples.
After that, we perform MCTS evolution on the seed data and update the policy model.
At the end of the second evolution, we further enhance the policy with DPO and train the PRM using the second iteration's data.

\begin{figure}[tbp]
    \centering
    \includegraphics[width=0.8\linewidth]{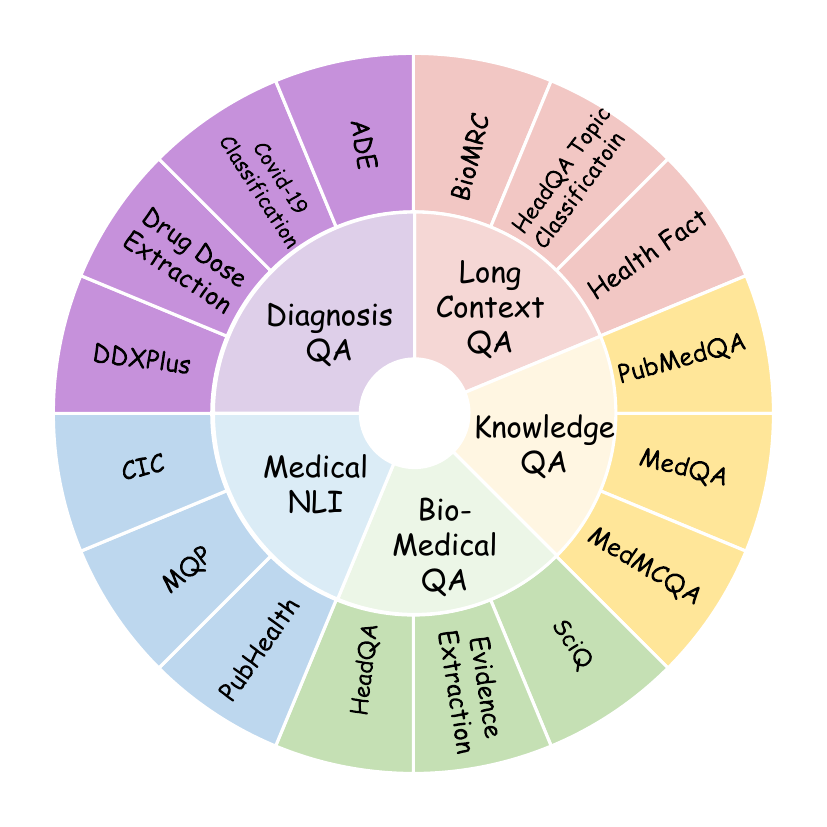}
    \caption{Overview of the used seed datasets.}
    \label{fig: seed_data_overview}

\end{figure}

\begin{table*}[tp]
\centering
\small
\setlength{\tabcolsep}{1mm}
\begin{tabular}{lccccc|cccccc|c}
    \toprule
\textbf{Models}
          & \textbf{MedQA} & \textbf{MedMCQA} & \textbf{PQA.} & \textbf{BioASQ} & \textbf{MMLU} & \textbf{BioMRC} & \textbf{PubH.} & \textbf{HFact.} & \textbf{DDX+}$^\dagger$ & \textbf{DrugD.}$^\dagger$ & \textbf{SEER}$^\dagger$ & \textbf{Avg.} \\
    \midrule
    
    \multicolumn{6}{l|}{\textit{\textbf{Large language models ($>$10B)}}}            &       &       &       &       &       &       &  \\
    GPT-4o-mini & 75.81 & 67.58 & 47.80 & 83.01 & 83.79 & 66.85 & 59.14 & 65.24 & 54.00 & 73.91 & 54.54 & 66.52 \\
    GPT-3.5-turbo & 59.31 & 58.12 & 37.40 & 74.11 & 71.11 & 56.22 & 57.84 & 67.85 & 39.05 & 86.96 & 73.61 & 61.96 \\
    QwQ-32B-preview & 68.89 & 61.03 & 48.60 & 73.62 & 74.18 & 79.76 & 63.36 & 66.08 & 45.40 & 39.13 & 37.26 & 59.76 \\
    R1-Distill-Qwen32B & 76.83 & 66.27 & 38.20 & 78.32 & 85.07 & 78.66 & 59.95 & 63.80 & 53.90 & 82.61 & 26.22 & 64.53 \\
    \midrule
    \multicolumn{6}{l|}{\textit{\textbf{Small language models ($<$10B)}}}             &       &       &       &       &       &       &  \\
    Qwen2.5-7B & 55.54 & 54.12 & 53.40 & 73.62 & 74.38 & 56.48 & 57.11 & 52.69 & 31.25 & 60.87 & 33.07 & 54.78 \\
    Llama3-8B & 57.50 & 55.92 & 56.40 & 75.73 & 68.55 & 56.50 & 64.09 & 70.88 & 35.30 & 73.91 & 47.07 & 60.17 \\
    Llama3.1-8B & 61.51 & 57.42 & 59.00 & 71.36 & 72.52 & 55.60 & 61.82 & 63.97 & 19.00 & 73.91 & 52.62 & 58.98 \\
    R1-Distill-Llama8B & 50.12 & 48.89 & 46.60 & 70.55 & 68.42 & 53.49 & 55.73 & 62.04 & 36.10 & 69.57 & 31.71 & 53.93 \\
    \midrule
    \multicolumn{6}{l|}{\textit{\textbf{Small Medical language models ($<$10B)}}}             &       &       &       &       &       &       &  \\
    MedLlama3 & 55.85 & 59.36 & 66.40 & \underline{84.63} & 70.08 & 47.97 & 62.39 & 68.10 & 22.50 & 69.57 & 50.69 & 59.78 \\
    Med42 & 50.20 & 49.70 & 55.40 & 74.76 & 61.43 & 57.26 & 59.14 & \textbf{81.57} & 31.35 & 65.22 & 37.14 & 56.65 \\
    OpenBioLLM & 50.20 & 50.56 & 41.40 & 47.73 & 61.69 & 27.46 & 18.77 & 53.28 & 16.55 & 34.78 & 46.48 & 40.81 \\
    UltraMedical3-8B & 68.89 & 61.82 & 51.60 & 80.58 & 75.08 & 45.18 & 66.13 & 72.73 & 36.70 & 60.87 & 24.55 & 58.56 \\
    UltraMedical3.1-8B & 70.93 & 62.78 & 56.40 & 77.18 & 76.43 & 54.26 & 59.14 & 70.20 & 31.55 & 56.52 & 45.86 & 60.11 \\
    m1-7B-32K & 70.70 & 61.85 & 48.60 & 77.83 & 78.35 & 52.93 & 56.70 & 61.62 & 29.15 & 69.57 & 56.70 & 60.36 \\
    HuatuoGPT-o1 & 62.53 & 59.31 & 58.20 & \textbf{87.70} & 70.53 & 50.98 & 24.61 & 66.08 & 40.20 & 56.52 & 46.85 & 56.68 \\
    \midrule
    \multicolumn{6}{l|}{\textbf{\mone (ours)}}             &       &       &       &       &       &       &  \\
    \quad Iter 1 & 65.91 & 60.55 & 56.80 & 78.48 & 75.66 & 55.84 & 57.03 & 64.73 & 51.65 & 73.91 & 48.97 & 62.68 \\
    \quad Iter 2 & 67.09 & 61.56 & 60.40 & 80.93 & 75.21 & 70.11 & 68.97 & 69.87 & 53.55 & \textbf{91.30} & 53.44 & 68.40 \\
    \quad Iter 2 \textit{w/ PRM} & \textbf{72.97} & \textbf{67.32} & \textbf{64.20} & 81.39 & \textbf{79.63} & \textbf{74.54} & \textbf{74.41} & \underline{76.18} & \textbf{62.40} & \textbf{91.30} & \textbf{59.80} & \textbf{73.10} \\
        \bottomrule
    \end{tabular}%

\caption{Experiment results in 11 in-domain datasets. We highlight the best results with \textbf{bold} and \underline{underlines} the second-best results among models with a similar size. `PQA.' denotes `PubMedQA', `PubH.' denotes `PubHealth', `HFact.' denotes `HealthFact', and `DrugD.' denotes `DrugDose'.  $^\dagger$ denotes that the ground truth is not a simple choice index. 
}
\label{tab:main_result}
\end{table*}

\section{Data Statistics}
\label{sec:dataset}
A slow-thinking system in medical scenarios should both excel at exam-level question answering~(QA) and handling real-world clinical scenarios, like diagnosis~\citep{Fansi_Tchango2022}, specific disease syndrome~\citep{covid19classification2020} and drug-related queries~\citep{huynh-etal-2016-adverse}.
However, previous works mainly focused on a simple scenario, with only limited data diversity, especially multiple-choice QA, to train reasoning models.
To approximate realistic clinical usage and promote medical reasoning models on a broader range of clinical tasks, we curate a training corpus from 16 existing public medical datasets and divide them into five dimensions according to the task category. 
The five dimensions, i.e., clinical diagnosis QA, natural language inference, knowledge-intensive QA, long-context QA, biomedical QA and corresponding datasets are shown in Fig.~\ref{fig: seed_data_overview}.

\section{Experiments}
\label{main_exp}
In this section, we comprehensively evaluate \mone on both in-domain and out-of-domain datasets.

\subsection{Experiment Setups}
\label{exp_setups}
\paragraph{Training and Evaluation}
We choose Llama3.1-8B-Instruct as the backbone of \mone.
We select MedQA-5op~\citep{jin2021disease}, PubMedQA~\citep{jin-etal-2019-pubmedqa} without contexts, MedMCQA~\citep{pal2022medmcqa}, PubHealth~\citep{kotonya-toni-2020-explainable-automated}, BioMRC~\citep{pappas-etal-2020-biomrc}, HealFact Classification~\citep{kotonya-toni-2020-explainable-automated}, Drug Dose Extraction~\citep{huynh-etal-2016-adverse}, DDXPlus~(DDX+; \citet{Fansi_Tchango2022}), the medical subsets of MMLU~\citep{hendrycks2021measuring}, BioASQ~\citep{tsatsaronis2012bioasq} SEER Classification~\citep{dubey2023using} as the evaluation sets.

\paragraph{Baselines}
We choose the following three categories to serve as baselines: (1) LLMs, including GPT-3.5-turbo~\citep{elmohamed}, GPT-4o-mini~\citep{gpt4}, QWQ-preview-32B~\citep{qwq-32b-preview} and R1-Distill-Qwen32B~\citep{guo2025deepseek}; (2) Small Language models ($<$10B), including Llama 3 8B, Llama 3.1 8B~\citep{DBLP:journals/corr/abs-2407-21783} and Qwen2.5 7B~\citep{qwen2}, R1-Distill-Llama8B~\citep{guo2025deepseek} (3) Medical LLMs, including MedLlama 3 8B~\citep{MedLLAMA3}, 
Med42~\citep{med42v2}, OpenBioLLM~\citep{OpenBioLLMs}, UltraMedical3-8B and UltraMedical3.1-8B~\citep{zhang2024ultramedical}, m1-7B-32K~\citep{huang2025m1UnleashPotential} and HuatuoGPT-o1-8B~\citep{chen2024huatuogpt}.
All the baselines are evaluated using CoT while \mone \textit{w/} PRM scores each response with the minimum step value and uses Best-of-N (N=32) to select the final response.

  
  
\begin{table}[tbp]
  \centering
  \setlength{\tabcolsep}{1mm}
  \small
    \begin{tabular}{lccc}
    \toprule
    \textbf{Model} & \textbf{MedCalc} & \textbf{MedXpert} & \textbf{RDC} \\
    \midrule
    GPT-4o-mini & 29.80 & 15.43 & 37.80 \\
    HuatuoGPT-o1 & 21.97 & 16.04 & 16.20 \\
    UltraMedical-3.1-8B & 15.19 & 16.12 & 21.80 \\
    R1-Distilled-Llama-8B & 11.94 & 12.65 & 13.60 \\
    \midrule
    \mone & 23.69 & 16.20 & 33.20 \\
    \mone \textit{w/ }PRM & \textbf{30.66} & \textbf{16.44} & \textbf{41.20} \\
    \bottomrule
    \end{tabular}%
  \caption{Out-of-domain comparison between \mone and previous state-of-the-art models. \mone achieves great generalization ability on both policy and process reward models.}
  \label{tab:ood_generalization}%
\end{table}%
\subsection{Main Results}
We present the experiment results in Table~\ref{tab:main_result}, splitting into examination QA and clinical application tasks.
The results unveil that most prior medical LLMs show superior results in traditional multiple-choice problems; while such superiority falls short on out-of-distribution real-world clinical benchmarks (DDXPlus or SEER), resulting in a sub-optimal overall performance compared to the general LLM--Llama3-8B.
In contrast, our \mone is tailored for universal medical applications and hence achieves the best overall performance among all open-sourced competitions.
As an 8B system, \mone achieves +14.12 average performance gains with respect to the base model in the overall assessment, outperforming both medical-oriented models and general reasoning models.
After two iterations, the policy model individually achieved the state-of-the-art (SoTA) performance, based on which the soft dual-side PRM further brings an additional 4.7 points improvement.
Notably, unlike previous methods that rely on large volumes of multiple-choice queries and consequently suffer from over-fitting, \mone achieves robust reasoning improvements, demonstrating that as few as 1,000 high-quality seed examples per task are sufficient to attain superior clinical reasoning performance.


\subsection{Generalization to Out-of-domain Tasks}
To validate the efficacy of \mone on real-world tasks with little labeled data, we select the most frontier models, including GPT-4o-mini, R1-Distill-Llama8B, HuatuoGPT-o1 and UltraMedical3.1-8B as the competitors and further compare \mone on MedCalc~\citep{khandekar2024medcalc}, MedXpert~\citep{zuo2025medxpertqa} and the rare disease confirmation (RDC) part sourced from PMCPatients~\citep{zhao2023alarge}.
Experiment results in Table~\ref{tab:ood_generalization} illustrate that both the policy and the PRM are applicable to unseen problems and the reasoning manner incentivized by self-evolution is sufficient for both clinical rare disease reasoning and more challenging reasoning scenarios.

\begin{table*}[tbp]
  \centering
  \small 
  \setlength{\tabcolsep}{1mm}
            \begin{tabular}{lcccccccccccc}
    \toprule
    \textbf{Setting} & \textbf{MedQA} & \textbf{MedMCQA} & \textbf{PQA.} & \textbf{BioASQ} & \textbf{MMLU} & \textbf{BioMRC} & \textbf{PubH.} & \textbf{HFact.} & \textbf{DDX+} & \textbf{DrugD.} & \textbf{SEER} & \textbf{Avg.} \\
    \midrule
    SFT Policy & 64.69 & 61.46 & 57.80 & 80.26 & 75.98 & 63.28 & 63.44 & 64.23 & 52.65 & 78.26 & 48.85 & 64.63 \\
     \quad\textit{w/} DPO & 67.09 & 61.56 & 60.40 & 80.93 & 75.21 & 70.11 & 68.97 & 69.87 & 53.55 & \textbf{91.30} & 53.44 & 68.40 \\
     \quad\textit{w/} H-S label & 68.97 & 65.67 & 61.80 & 79.45 & 76.75 & 70.48 & 69.13 & 74.24 & 59.35 & 86.96 & 56.94 & 69.98 \\
     \quad\textit{w/} H-D label & 66.77 & 63.78 & 61.40 & 80.74 & 75.14 & \textbf{78.13} & 69.54 & 75.34 & 61.60 & \textbf{91.30} & 56.46 & 70.93 \\
     \quad\textit{w/} S-D label & \textbf{72.97} & \textbf{67.32} & \textbf{64.20} & \textbf{81.39} & \textbf{79.63} & 74.54 & \textbf{74.41} & \textbf{76.18} & \textbf{62.40} & \textbf{91.30} & \textbf{59.80} & \textbf{73.10} \\
     \quad\textit{w/} SFT init. PRM & 70.70 & 64.40 & 61.80 & 81.23 & 77.39 & 70.22 & 75.30 & 74.58 & 60.15 & 82.61 & 54.99 & 70.31 \\

    \bottomrule
    \end{tabular}%

  \caption{\blue{Ablation study on each component of \mone after the second iteration. ``H-S'' means hard single-sided label, ``H-D'' means hard dual-sided label, and ``S-D'' is soft dual-sided label used in \mone. }}
  \label{tab:ablation}%
\end{table*}%

\section{Analysis}

\subsection{Ablation Study}
In this section, we validate the effectiveness of each sub-module of \mone.
Starting from the SFT-tuned policy model, we compare the final performance with (1) \textit{w/} DPO: use DPO to fine-tune the policy; (2) \textit{w/} H-S label: conduct best-of-N evaluation using a PRM trained with hard single-sided label~\citep{zhang2025lessons}; (3) \textit{w/} H-D label: same as (2) but use hard dual-sided label~\citep{wang2025towards} to train a PRM and (4) \textit{w/} S-D label (ours): same as (2) but use soft dual-sided label proposed in \mone to train a PRM.
We also compare with (5) \textit{w/} SFT init. PRM, which is the same as (4) but initializes PRM from the SFT-tuned policy, to further show the significance of a PRM exposed to both positive and negative responses.
Experiment results in Table~\ref{tab:ablation} show that the DPO helps to greatly improve the policy model, especially in clinical tasks.
Furthermore, innovatively determining the dual side label based on the MC estimation, our method is more robust and flexible than rule-based labels, and hence outperforms previous training objectives, confirming the necessity of holistic modeling of a PRM.
\begin{figure}[tbp]
    \centering
    \includegraphics[width=0.9\linewidth]{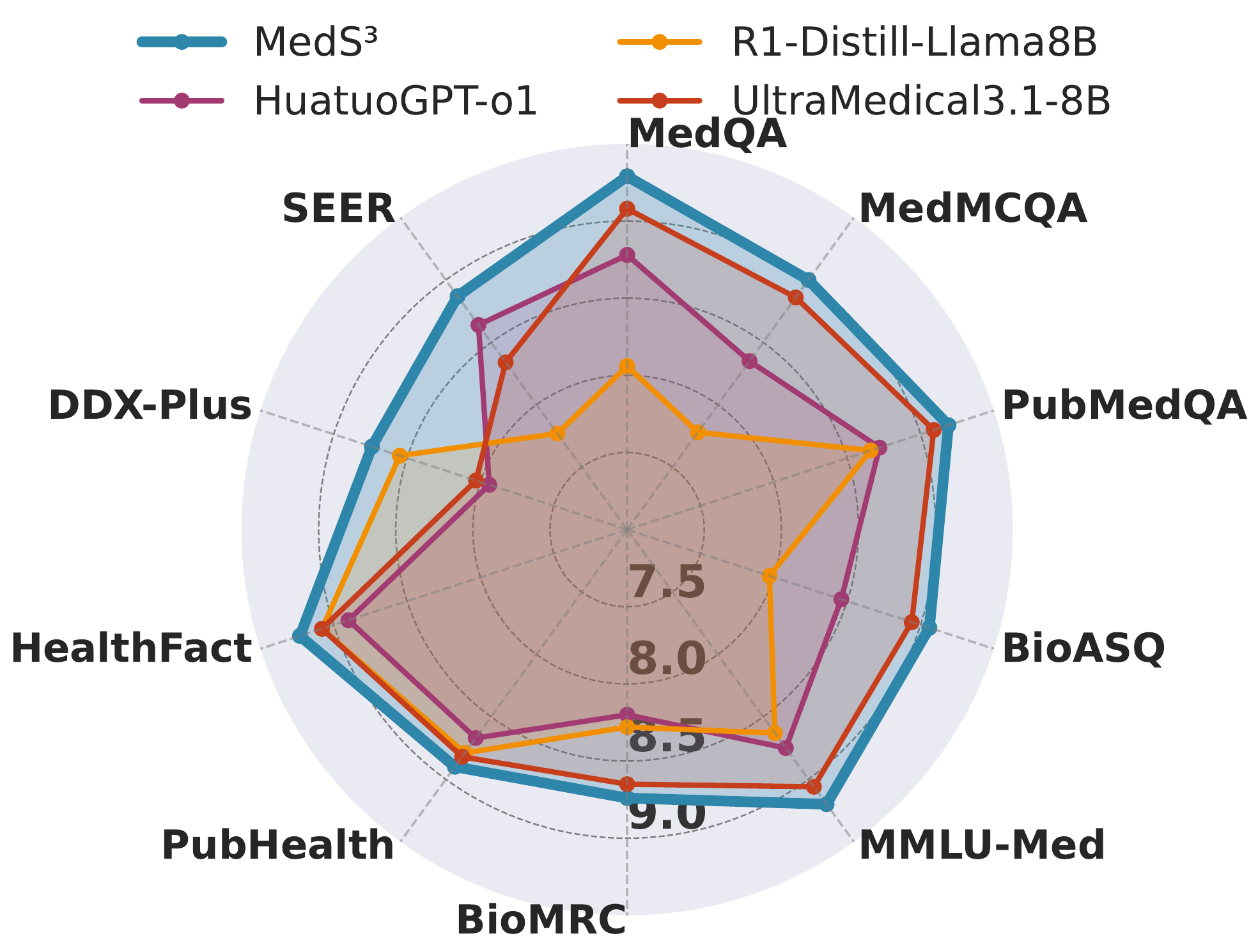}
    \caption{Interpretability evaluation for models using synthetic data, where \mone produces the least hallucinatory contents among other pioneering models.}
    \label{fig: gpt_eval}

\end{figure}


\subsection{Reliability of \mone}
The non-eliminable hallucinations prevent most medical LLMs from being practical.
Albeit inevitability, \mone leverages a fine-grained soft dual-sided PRM to improve interpretability and mitigate hallucinatory contents.
We leverage GPT-4o to evaluate baselines that rely on fine-tuning on synthetic datasets, including HuotuoGPT-o1, R1-Distilled-Llama-8B and UltraMedical3.1-8B, where each model's output is scored based on its medical reasonableness, logical coherence, and explainability.
DrugDose is excluded from evaluation due to its small size and consequently unreliable statistical significance. 
The evaluation prompt is presented in Appendix E. 
Results in Fig.~\ref{fig: gpt_eval} indicates that \mone achieves the highest evaluation score.
We attribute such superiority to the dual awareness of the PRM, which is trained to penalize wrong intermediate steps, and therefore could induce a relatively lower score to trajectories containing hallucinations and a correct final answer.
The Best-of-N strategy avoids picking up such trajectories and enhances the interpretability of the final output.

\begin{figure}[tbp]
    \centering
    \begin{subfigure}{0.32\linewidth}
        \centering
        \includegraphics[width=\linewidth]{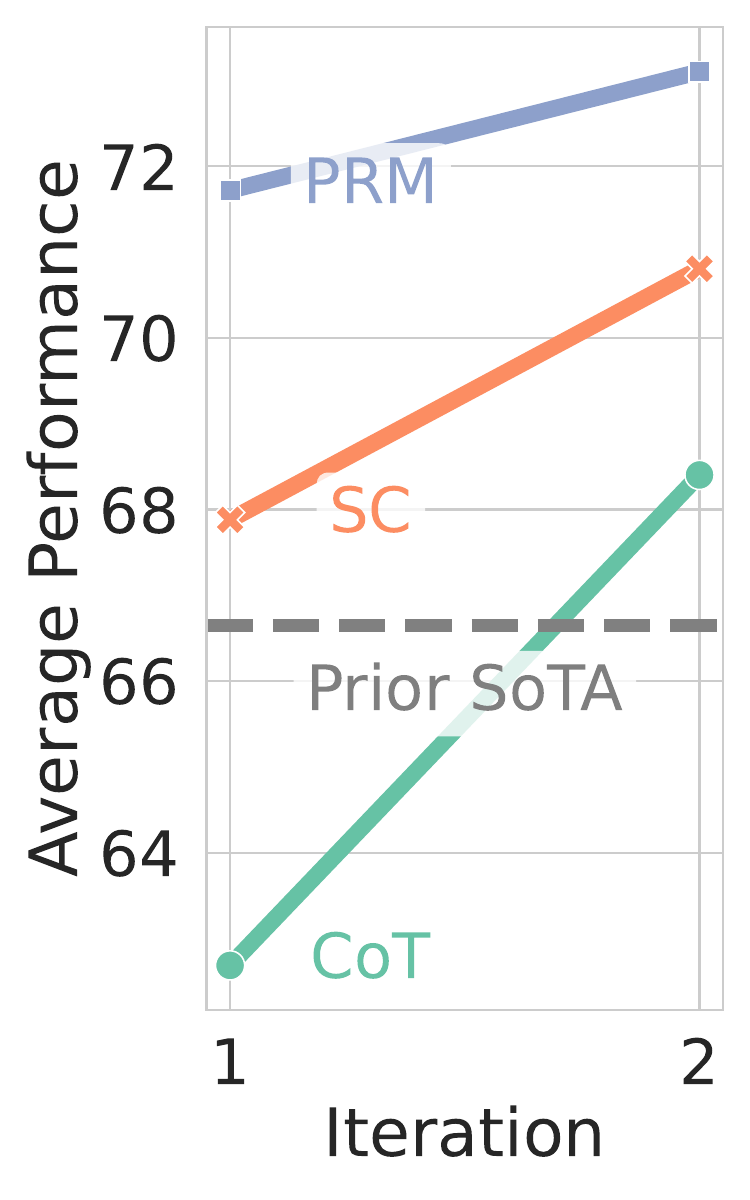}
        \caption{}
        \label{fig:iteration}
    \end{subfigure}%
    \begin{subfigure}{0.64\linewidth}
        \centering
        \includegraphics[width=\linewidth]{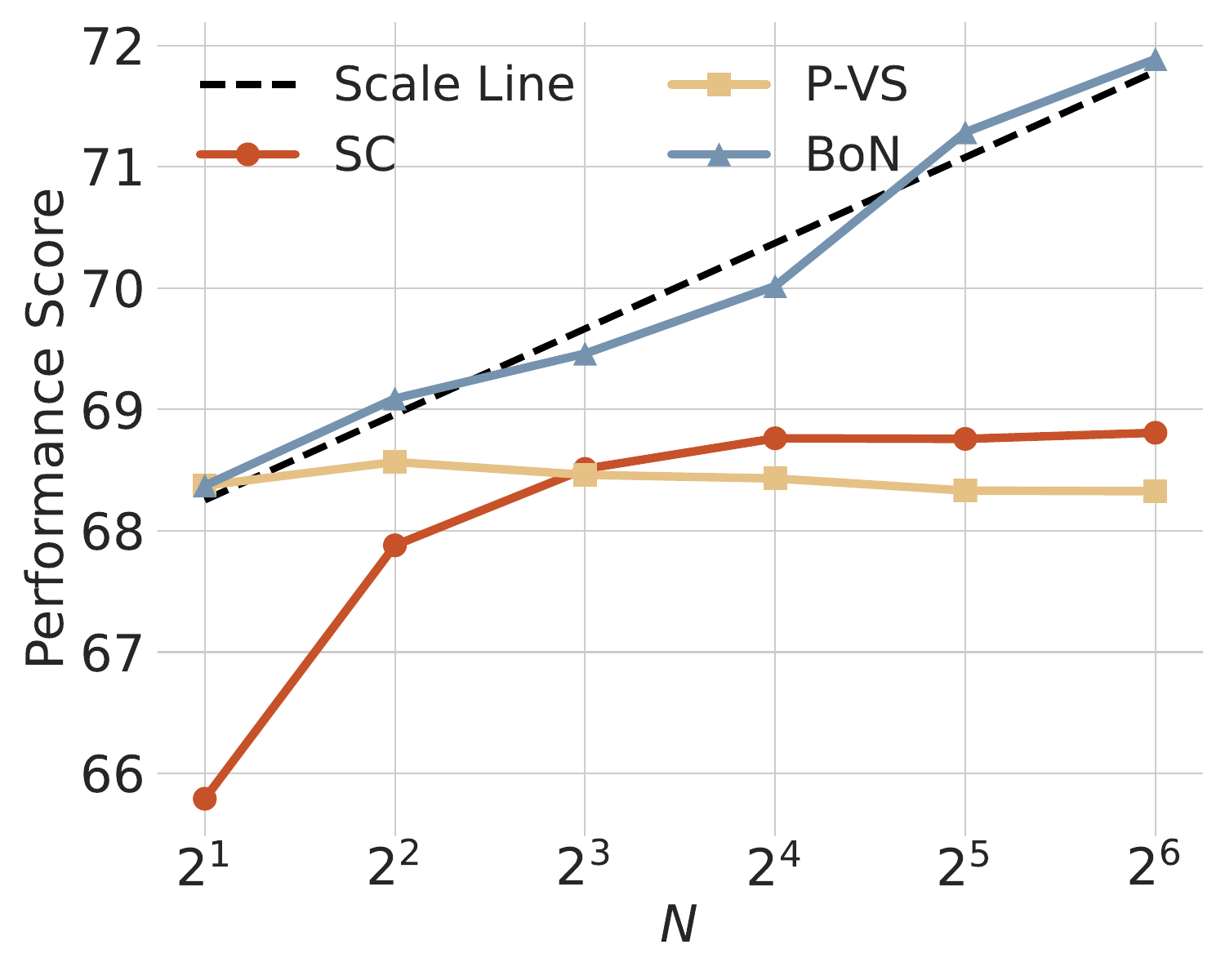}
        \caption{}
        \label{fig: test_time_scale}
    \end{subfigure}
    \caption{Scaling in (a) self-evolution iterations and (b) sampling numbers during test-time. Both the policy and PRM harvest consistent enhancement with self-evolution, and hence their cooperative system \mone achieves a log-linear scaling rate with little saturation.  }
    \label{fig: scaling}
\end{figure}

\subsection{Scaling of \mone}
In this section, we present the improvements brought by the self-evolutionary framework in Fig.~\ref{fig:iteration}, and those attributable to test-time scaling in Fig.~\ref{fig: test_time_scale}.
Specifically, we sample $n=2,4,8,16,32,64$ candidates for a prompt with a 1.0 temperature and compare the performance obtained through Best-of-N (BoN)~\citep{lightman2023let}, PRM-guided Vote-Sum~(P-VS; \citet{wang-etal-2024-math}), as well as an SC baseline.
We observe a great improvement in both the policy model and the PRM after a second evolution iteration, highlighting the efficacy of self-evolution.
This suggests that the iterative MCTS process, where the model learns from its own refined outputs, leads to steadily increased improvements.
Additionally, we find that test-time scaling further enhances \mone's reasoning performance as illustrated in Fig.~\ref{fig: test_time_scale} in an effective log-linear rate with little saturation. 
Together, these results highlight the benefits of both self-exploration during synthesis and self-supervision during inference, contributing to \mone's strong performance across diverse tasks.
Note that the P-VS performance is inferior to BoN, as most plausible reasoning chains arriving at correct answers but with incorrect reasoning steps are labeled with low values by our soft dual-sided PRM. 
Although these hallucinatory chains deteriorate the grouping of correct answers, our PRM still could assign the highest score for trajectories with both correct reasoning steps and final answers, therefore contributing to a log-linear scaling on BoN performance.

\begin{figure}[tbp]
    \centering
    \includegraphics[width=0.9\linewidth]{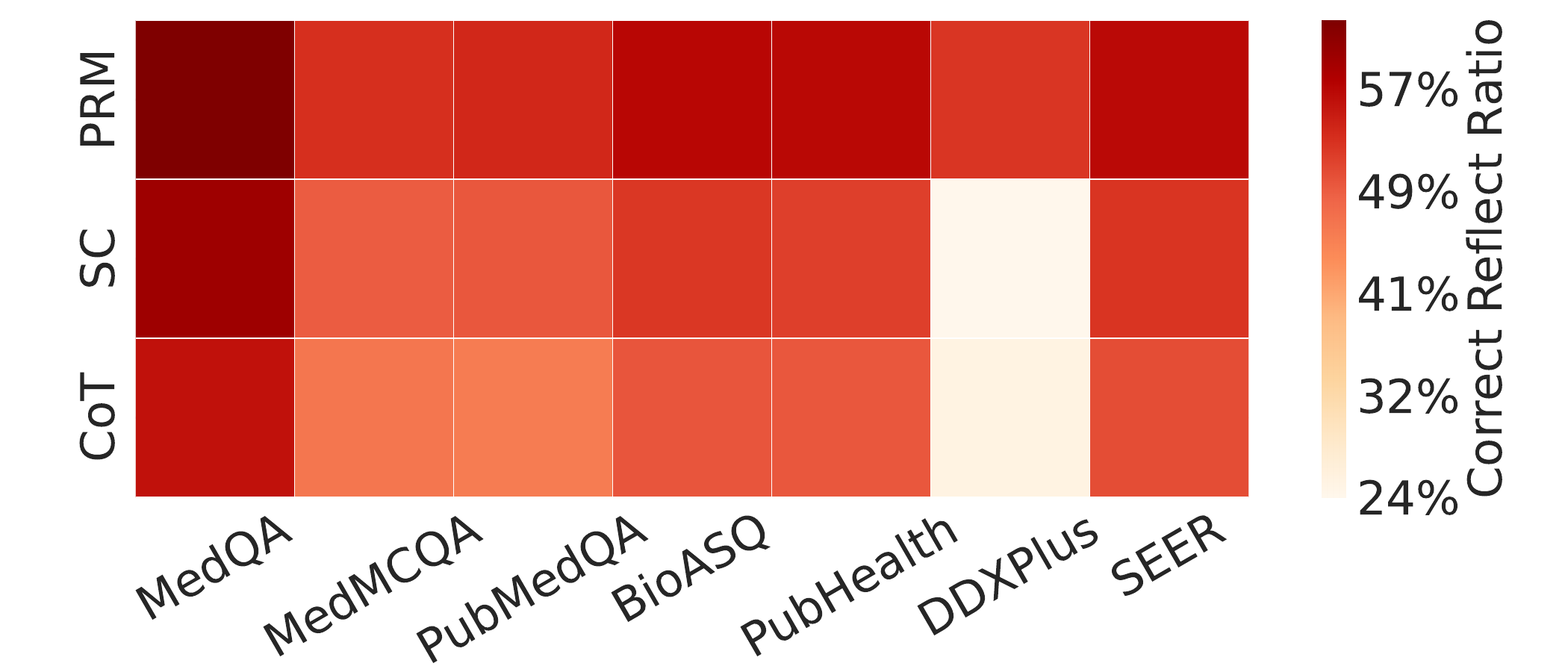}
    \caption{Reflective response ratio of \mone across 7 representative datasets. Both the policy and PRM are reflection-aware to perform sequential test-time scaling.}
    \label{fig: introspect_rate}
\end{figure}

\subsection{Introspective Behavior}
Reflection has been proved to be an effective scaling paradigm for enhancing LLM's test-time scaling capacity~\citep{guo2025deepseek}.
Our \mone introduced a \texttt{Reflect} node during synthesis and a soft dual-sided PRM to encourage correctly reflected responses, aiming to impart self-reflection behavior to the whole system.
We manually define reflective tokens (\texttt{Wait, reevaluate, recheck, however, but}) and count the ratio of correct responses with these tokens on seven representative benchmarks in Fig.~\ref{fig: introspect_rate}.
We observe a steady increase in the occurring ratio from directly chain-of-thought prompting to leveraging PRM to conduct BoN evaluation, which indicates both the policy and PRM in \mone has been imparted with self-reflection behavior.
This further demonstrates that the PRM trained with the soft dual-sided label can correctly favor valuable responses with self-reflection.



\subsection{Comparison of Reasoning Styles}
\begin{figure}[tbp]
    \centering
    \includegraphics[width=0.9\linewidth]{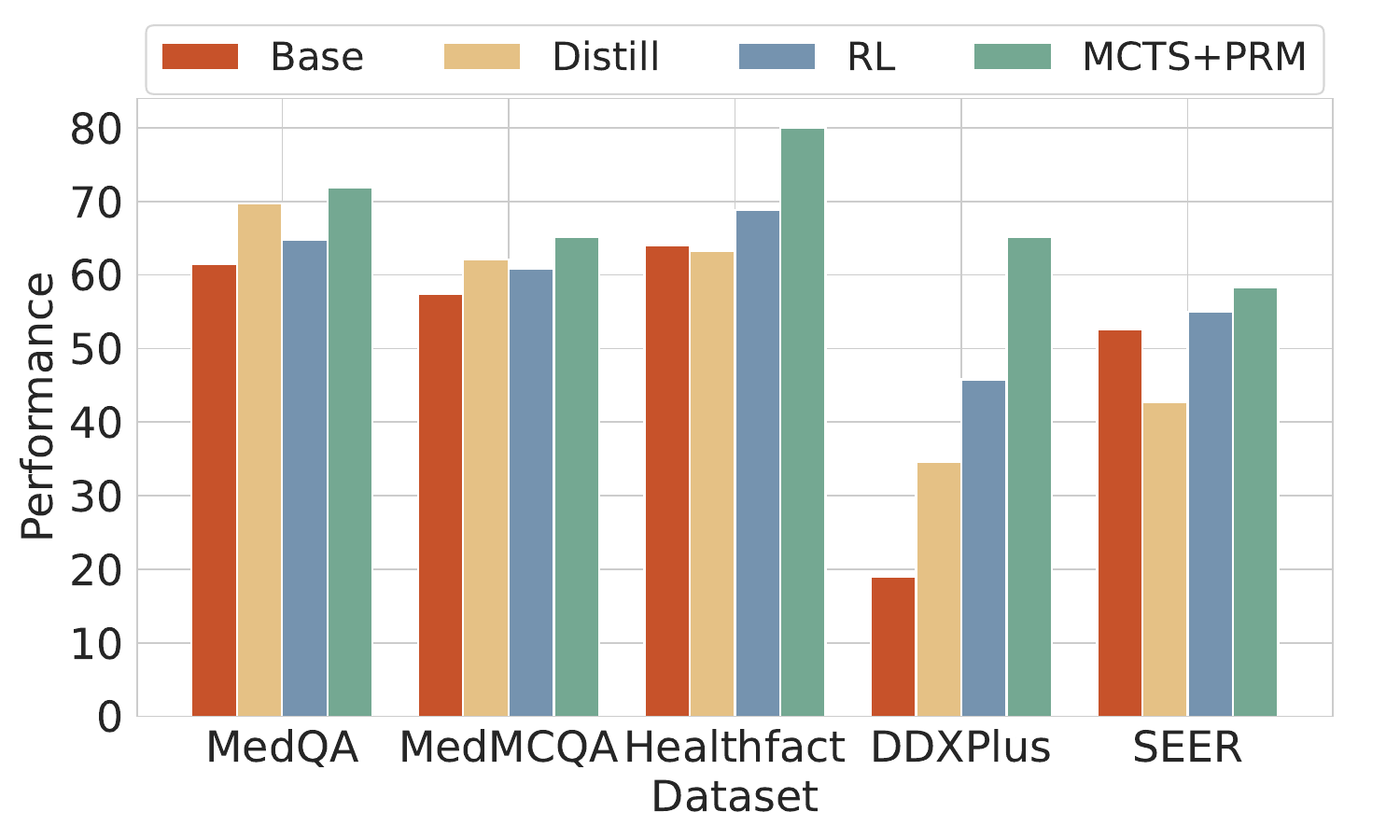}
    \caption{Three widely adopted methods to empower models with medical reasoning abilities. MCTS+PRM is the best among the three, making it the core of \mone. }
    \label{fig:methods}
\end{figure}

In this section, we compare three reasoning enhancement strategies, including MCTS plus PRM which is what \mone leverages, with distillation from strong reasoning models, which is what O1-journey-part3~\citep{huang2025o1} does and pure reinforcement learning~(RL), which is what DeepSeek-R1~\citep{guo2025deepseek} adopts.
We use the first iteration dataset in \S\ref{sec:dataset} to implement RL, and use the officially released distillation dataset provided by \citet{huang2025o1} to SFT the base model, and compare them with \mone after the first evolution iteration. 
The results presented in Fig.~\ref{fig:methods} demonstrate that in exam-level medical QA datasets where the base model already excels at, distillation from large proprietary reasoning models is much more data-efficient than the other two methods, albeit sacrificing generalization in clinical tasks.
In contrast, with both a considerable performance leap and generalization, RL is second to MCTS+PRM.
We hypothesize that the soundness of medical diagnosis step is clear to determine, reducing reward hacking and resulting in a more reliable PRM and credible preference estimation.


\section{Related Works}

\paragraph{Slow-Thinking Medical LLMs} With the significant achievements of the o1~\citep{jaech2024openai} in complex reasoning tasks, previous works show the potential advantage of the o1-like models in medical tasks~\citep{xie2024preliminary,nori2024medprompt}. Based on these, previous works develop the slow-thinking medical LLMs with distillation: \citet{huang2025o1} directly learn the reasoning trajectory generated by o1 and \citet{chen2024huatuogpt} improve the model’s reasoning ability through o1 synthesis of reflective data and reinforcement learning. 
Besides, \citet{yu2025finemedlm} create a Chinese version slow-thinking medical LLMs by constructing the preference data with QwQ~\citep{qwq-32b-preview}.

\paragraph{Self-Evolving Reasoning and Process Supervision}
Recent work in self-improving reasoning systems has explored Monte Carlo Tree Search (MCTS)~\citep{zhang2024rest} and reinforcement learning to enable models to refine their own outputs~\citep{guo2025deepseek}. 
Methods like Tree of Thoughts (ToT)~\citep{yao2023tree} demonstrate the potential of search-based exploration for generating high-quality reasoning trajectories. 
Concurrently, process reward models (PRMs) have been proposed to provide step-wise feedback~\citep{lightman2023let}, yet most assume binary correctness based on potential correctness and fail to penalize reasoning degradation. 

\section{Conclusion}
In this paper, we present \mone, a self-evolved slow-thinking system built for universal clinical usage.
We extend the clinical reasoning to diverse tasks to enhance generalization, and 
use MCTS to construct policy data and PRM data.
We propose a new PRM learning objective -- the soft dual-sided label, which enables the PRM to reward a step based on both future and past aspects, to produce credible long-chain reflective responses.
Experiment results demonstrate that \mone achieves superior performance on diverse medical benchmarks, especially in realistic clinical ones, surpassing open-sourced models by a large margin with fewer parameters.

\section*{Acknowledgments}
This work is supported by National Key R\&D Program of China (No. 2022ZD0162101), National Natural Science Foundation of China (No. 62576209) and STCSM (No. 2025SHZDZX025G05).

\bibliography{aaai2026}
\clearpage
\appendix

\appendix
\begin{figure}[tbp]
    \centering
    \includegraphics[width=0.95\linewidth]{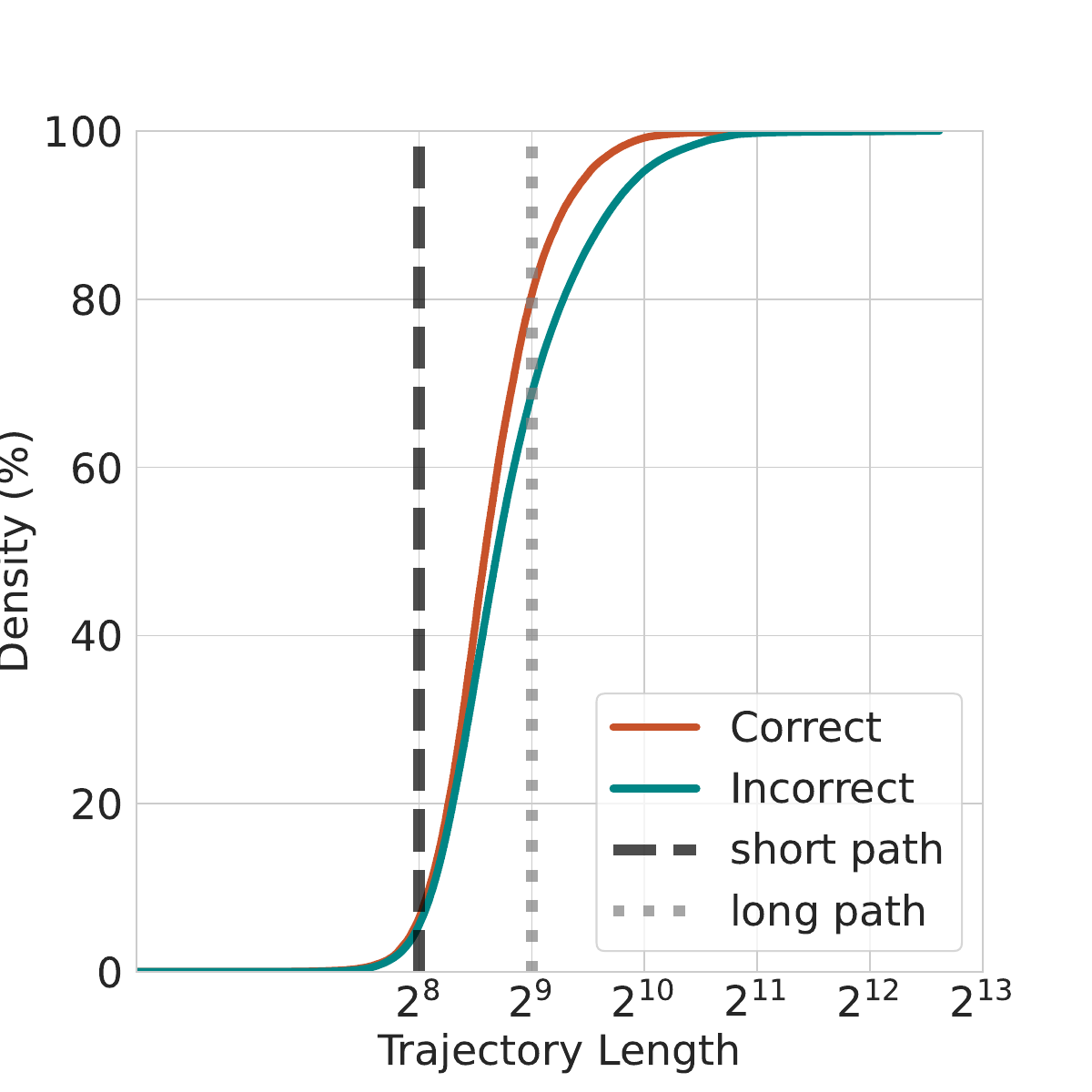}
    \caption{Trajectory length distribution of correct and incorrect sets of the evolved dataset. }
    \label{fig: statistics}
\end{figure}

\blue{
\section{Statistics of the Evolved Dataset}
\label{sec:dataset_statistics}
In this section, we show the statistics of the evolved dataset after the second evolution, which is used to conduct our training of \mone. 
We plot the length distributions of trajectories in Fig.~\ref{fig: statistics}.
Defining short sequences as below 256 and long sequences as above 512, we find that the evolved dataset contains about 20\% long trajectories, which enables the policy model to generate reliable responses with more tokens.
Moreover, we find that correct trajectories consume fewer tokens than incorrect ones, which aligns highly with \citet{zeng2025Revisiting}.
}

\section{Limitations}
\mone achieves superior performance over eleven benchmarks by conducting MCTS in seed datasets to collect both policy and PRM training data and a newly proposed PRM learning objective: soft dual side label.
However, it can be further improved via these strategies: (1) adopt human-in-the-loop strategies to further enhance the interpretability of medical reasoning LLMs; (2) introduce more training samples to cover more medical reasoning scenarios; (3) conduct more evolution iterations to further improve the model.


\section{Further Experiments}
In this section, we present more experiments to validate the effectiveness of \mone.

\begin{table}[tbp]
  \centering
    \begin{tabular}{lcccc}
    \toprule
    $\beta$ & 0.5   & 1     & 1.5   & 2 \\
    Dev loss & 0.4293 & 0.4169 & 0.4194 & 0.538 \\
    \bottomrule
    \end{tabular}%
  \caption{Grid search of $\beta$ and corresponding loss in the dev set.}
  \label{tab:beta_search}%
\end{table}%

\subsection{Determination of $\beta$}
We perform a simple grid search on a pre-defined dev set to find the most appropriate value o f $\beta$ in Eq.~\ref{eq:soft_dual_side_label}.
Specifically, we search $\beta$ in the following list $[0.5,1.0,1.5,2.0]$ and show the loss in the development set in Table~\ref{tab:beta_search}.
We determine $\beta$ as 1 for its lowest loss.
Although there might exist a more advanced configuration, we just set $\beta$ to 1 as this is not our focus and we leave this for future work.




\begin{table*}[tbp]
\setblue
  \centering
  \resizebox{\textwidth}{!}{%
    \begin{tabular}{lcccccccccccc}
    \toprule
    \textbf{Model} & \textbf{\newline{}MedQA} & \textbf{MedMCQA} & \textbf{PubmedQA} & \textbf{Bioasq} & \textbf{Med MMLU} & \textbf{Biomrc} & \textbf{Pubhealth} & \textbf{Healthfact} & \textbf{DDX Plus} & \textbf{Drug Dose} & \textbf{SEER} & \textbf{Average} \\
    \midrule
    Ultramedical3.1-8B & 72.90 & 65.57 & 58.00 & 80.58 & 78.54 & 49.18 & 67.91 & 71.72 & 34.10 & 60.87 & 49.73 & 62.65 \\
    m1-7b-32K & 72.38 & 64.55 & 50.40 & 80.91 & \textbf{80.19} & 59.50 & 58.08 & 68.10 & 31.80 & 82.61 & 50.91 & 63.58 \\
    HuatuoGPT-o1 & 72.86 & 66.94 & 58.20 & 78.48 & 78.54 & 46.45 & 64.58 & 70.29 & 41.00 & 60.87 & 49.59 & 62.81 \\
    \mone & \textbf{72.97} & \textbf{67.32} & \textbf{64.20} & \textbf{81.39} & 79.63 & \textbf{74.54} & \textbf{74.41} & \textbf{76.18} & \textbf{62.40} & \textbf{91.30} & \textbf{59.80} & \textbf{73.10} \\
    \bottomrule
    \end{tabular}%
    }
      \caption{\blue{Comparison with prior Top-2 models with similar model sizes prompted with self-consistency method. Albeit certain improvements, these models still lag behind \mone by a large margin.}}
  \label{tab:sc_cmp}%
\end{table*}%


\subsection{Comparison with SC Models}
\label{sec:sc_cmp}
\blue{
We also compare \mone with baselines prompted with the Self-Consistency (SC) method, which is a simple yet efficient way to scale in a parallel manner.
To maintain similar inference costs, we set the sampling number to 32 for models with similar size ($<$10B) and compare with the most powerful models before\footnote{It is reasonable when \mone outperforms the most leading baselines.}, namely HuatuoGPT-o1 and UltraMedical3.1-8B.
Results in Table~\ref{tab:sc_cmp} illustrate that as a test-time scaling method, SC improves the already strong baselines by significant gains, while such improvements usually occur in traditional benchmarks like MedQA or MedMCQA. 
Their performances in clinical testbeds, like SEER or DDX-Plus, hardly show gains, which unveils some kind of overfitting problem.
Therefore, optimized for both traditional examination and clinical usage, \mone robustly achieves state-of-the-art performance overall.
}

\begin{table*}[tbp]
\setblue
  \centering
  \resizebox{\textwidth}{!}{%
        \begin{tabular}{lcccccccccccc}
    \toprule
    \textbf{Backbone} & \textbf{MedQA} & \textbf{MedMCQA} & \textbf{PubMedQA} & \textbf{BioASQ} & \textbf{Med MMLU} & \textbf{BioMRC} & \textbf{PubHealth} & \textbf{HealthFact} & \textbf{DDX Plus} & \textbf{Drug Dose} & \textbf{SEER} & \textbf{Average} \\
    \midrule
    UltraMedical3.1 & \textbf{68.42} & 58.20 & 58.00 & \textbf{79.61} & 73.16 & 49.40 & 68.07 & 71.38 & 49.20 & 86.96 & 51.40 & 64.89 \\
    Llama 3.1 & 67.64 & \textbf{62.00} & \textbf{59.60} & 79.13 & \textbf{77.77} & \textbf{76.96} & \textbf{73.19} & \textbf{78.37} & \textbf{63.80} & \textbf{91.30} & \textbf{59.20} & \textbf{71.72} \\
    \bottomrule

    \end{tabular}%
    }
  \caption{\blue{Comparison with UltraMedical-3.1-8B as the policy model. With superior instruction following ability and comparable medical knowledge, Llama-3.1-8B suits \mone system to fulfill the self-evolution procedure.}}
  \label{tab:backbone_cmp}%
\end{table*}%

\subsection{Backbone Selection}
\blue{
In this section, we investigate which backbone, a general LLM or a medical-specific LLM, is suitable for conducting self-evolution.
We conduct one iteration of evolution using the same data of \mone under UltraMedical3.1-8B and compare it with \mone after the \textbf{first iteration} using the Best-of-N decoding method to save computational cost. 
The comparison shown in the upper half of Table~\ref{tab:backbone_cmp} reveals that although UltraMedical3.1-8B improves compared to the initial policy, it still lags behind \mone by a large margin.
Delving into the generation, we find that UltraMedical3.1-8B suffers from endless generation, which stems from its lower instruction following ability compared to Llama 3.1 8B.
On the other hand, medical backbones show no significant performance gains compared to the general model (UltraMedical3.1 8B 60.11 vs Llama 3.1 8B 58.98 in Table~\ref{tab:main_result}), while after optimized for certain benchmarks like MedQA, they have lower generalization ability than general models.
Based on the above observations, we choose to use a general backbone with sufficient medical knowledge, i.e., Llama 3.1 8B, as the initial policy model.
}

\begin{table*}[tbp]
  \centering
  \resizebox{\textwidth}{!}{%
    \begin{tabular}{cccccccccccccc}
    \toprule
    \multicolumn{1}{l}{\textbf{Method}} & \textbf{Iteration} & \textbf{MedQA} & \textbf{MedMCQA} & \textbf{PubmedQA} & \textbf{Bioasq} & \textbf{Med MMLU} & \textbf{Biomrc} & \textbf{Pubhealth} & \textbf{Healthfact} & \textbf{DDX Plus} & \textbf{Drug Dose} & \textbf{SEER} & \textbf{Average} \\
    \midrule
    \multirow{6}[2]{*}{BoN} & 2     & 68.97 & 64.04 & 62.00 & 79.45 & 76.43 & 73.68 & 72.14 & 73.57 & 58.00 & 86.96 & 55.45 & 70.06 \\
          & 4     & 69.60 & 64.55 & 61.60 & 80.42 & 77.26 & 74.64 & 74.17 & 73.06 & 58.85 & 86.96 & 56.75 & 70.71 \\
          & 8     & 70.54 & 64.57 & 62.60 & 81.07 & 77.83 & 74.75 & 74.09 & 72.64 & 59.90 & 86.96 & 56.61 & 71.05 \\
          & 16    & 70.23 & 66.32 & 64.00 & 81.23 & 78.41 & 74.80 & 73.68 & 72.05 & 61.00 & 86.96 & 58.44 & 71.56 \\
          & 32    & 72.97 & 67.32 & 64.20 & 81.39 & 79.63 & 74.54 & 74.41 & 76.18 & \textbf{62.40} & 91.30 & 59.80 & 73.10 \\
          & 64    & \textbf{73.37} & \textbf{67.65} & \textbf{66.00} & \textbf{81.72} & \textbf{79.37} & 74.54 & \textbf{74.90} & \textbf{78.28} & 62.25 & \textbf{91.30} & \textbf{60.79} & \textbf{73.65} \\
    \midrule
    \multirow{6}[2]{*}{SC} & 2     & 65.67 & 61.49 & 60.60 & 77.02 & 73.73 & 71.09 & 68.48 & 70.79 & 56.45 & 91.30 & 52.59 & 68.11 \\
          & 4     & 67.09 & 63.11 & 60.40 & 78.80 & 75.72 & 73.23 & 70.59 & 76.18 & 57.35 & 91.30 & 56.32 & 70.01 \\
          & 8     & 67.40 & 63.71 & 60.60 & 80.42 & 76.30 & 73.82 & 70.11 & 77.61 & 57.65 & 91.30 & 57.48 & 70.58 \\
          & 16    & 68.42 & 63.73 & 60.80 & 80.42 & 76.43 & 73.70 & 70.11 & 77.69 & 58.05 & 91.30 & 58.27 & 70.81 \\
          & 32    & 67.64 & 63.52 & 60.60 & 80.26 & 76.55 & 73.98 & 70.59 & \textbf{78.28} & 57.90 & 91.30 & 58.25 & 70.81 \\
          & 64    & 67.79 & 63.45 & 60.80 & 80.26 & 76.75 & 73.98 & 70.76 & 77.86 & 58.10 & 91.30 & 58.33 & 70.85 \\
    \midrule
    \multirow{6}[2]{*}{P-VS} & 2     & 68.97 & 64.04 & 62.00 & 79.45 & 76.43 & 73.68 & 72.14 & 73.57 & 58.00 & 86.96 & 55.45 & 70.06 \\
          & 4     & 68.97 & 63.85 & 60.40 & 80.26 & 75.91 & 74.38 & 71.57 & 75.25 & 57.70 & 86.96 & 57.37 & 70.24 \\
          & 8     & 68.34 & 63.95 & 61.00 & 80.74 & 76.55 & 74.54 & 70.11 & 76.60 & 55.40 & 86.96 & 57.37 & 70.14 \\
          & 16    & 68.81 & 63.88 & 60.80 & 81.39 & 77.07 & 74.99 & 70.27 & 75.67 & 53.10 & 86.96 & 58.33 & 70.12 \\
          & 32    & 68.66 & 63.81 & 61.20 & 80.74 & 76.81 & 74.88 & 71.16 & 74.41 & 53.65 & 82.61 & 57.99 & 69.63 \\
          & 64    & 68.19 & 63.71 & 61.00 & 80.58 & 77.39 & \textbf{74.88} & 71.24 & 74.41 & 53.70 & 82.61 & 58.16 & 69.62 \\
    \bottomrule
    \end{tabular}%
    }
      \caption{Full table of test-time scaling using PRM with different evaluation methods.}
  \label{tab:scaling_full_table}%
\end{table*}%

\section{Future Work}
As a pioneering work, we have validated that small language models can self-evolve to empower themselves with strong reasoning abilities in clinical usage.
There are several remaining directions to further enhance \mone:
\begin{enumerate}
    \item Conduct Human-interference evaluation. MC-rollout value is verified to be not the best choice for evaluating the value of an internal step. We are eager to introduce a more fine-grained step label to enhance the optimization of the PRM.
    \item Introduce more clinical data, not limited to close-ended generation. Currently, all the data used in \mone are close-ended, and the application of reasoning is not limited to such a narrow room. We intend to extend \mone to broader clinical tasks to make \mone a more useful system.
\end{enumerate}
We will continue our exploration and make \mone more practical in medical domains.

\section{Prompt Template}

\label{sec:prompt_template}
We show the prompt used to synthesize reasoning data in Fig.~\ref{fig:reason_prompt}, Fig.~\ref{fig:finish_prompt}, and Fig.~\ref{fig:reflect_prompt}.
The prompt template to evaluate the hallucination of each medical LLM using synthetic data is shown in Fig~\ref{fig:evaluate_prompt}.

\begin{figure*}[tbp]
    \begin{promptbox}{Reason Template}
<|begin\_of\_text|><|start\_header\_id|>system<|end\_header\_id|>\newline

Cutting Knowledge Date: December 2023
Today Date: 23 July 2024\newline

<|eot\_id|><|start\_header\_id|>user<|end\_header\_id|>\newline

Reasoning Example:
{\{Few-shot Example\}}\newline

You are a professional medical expert majored at reasoning in hard medical-related problems.\newline

Think critically about the problem and answer with concise, accurate reasoning. Please ensure your reasoning is thorough and elaborate, breaking down each step of your thought process.\newline

Problem: \{problem\}<|eot\_id|><|start\_header\_id|>assistant<|end\_header\_id|>\newline

Step 0: Let's break down this problem step by step\newline

Step 1: 
\end{promptbox}
\caption{Reason template}
\label{fig:reason_prompt}

\end{figure*}

\begin{figure*}[tbp]
    \begin{promptbox}{Finish Template}
<|begin\_of\_text|><|start\_header\_id|>system<|end\_header\_id|>\newline

Cutting Knowledge Date: December 2023
Today Date: 23 July 2024\newline

<|eot\_id|><|start\_header\_id|>user<|end\_header\_id|>\newline

Reasoning Example:
{\{Few-shot Example\}}\newline

You are a professional medical expert majored at reasoning in hard medical-related problems.\newline

Use thorough and elaborate steps to complete your reasoning. Conclude the task by stating: "The answer is \{answer\}".\newline

Problem: \{problem\}<|eot\_id|><|start\_header\_id|>assistant<|end\_header\_id|>\newline

Step 0: Let's break down this problem step by step\newline

Step 1: 
\end{promptbox}
\caption{Finish template}
\label{fig:finish_prompt}

\end{figure*}

\begin{figure*}[tbp]
    \begin{promptbox}{Reflect Template}
<|begin\_of\_text|><|start\_header\_id|>system<|end\_header\_id|>\newline

Cutting Knowledge Date: December 2023
Today Date: 23 July 2024\newline

<|eot\_id|><|start\_header\_id|>user<|end\_header\_id|>\newline

Reasoning Example:
{\{Few-shot Example\}}\newline

You are a professional medical expert majored at reasoning in hard medical-related problems.\newline

Use thorough and elaborate steps to complete your reasoning. Conclude the task by stating: "The answer is \{answer\}".\newline

Problem: \{problem\}<|eot\_id|><|start\_header\_id|>assistant<|end\_header\_id|>\newline

Step 0: Let's break down this problem step by step\newline

Step 1: [omitted]\newline

Step k: [omitted]. The answer is C.\newline

Step k+1: Wait, the previous answer maybe incorrect and I need to reconsider other options.
\end{promptbox}
\caption{Reflect template}
\label{fig:reflect_prompt}

\end{figure*}

\begin{figure*}[tbp]
    \begin{promptbox}{Evaluate Template}
Given a medical problem and a response from a large language model (whose final prediction is correct), please based on the following criteria to give a score:\newline

Reasonableness: The response should be reasonable and consistent with the medical problem.\newline
Coherence: The response should be coherent and logically consistent.\newline
Explainability: The response should be explainable and easy to understand.\newline

Please give a score from 0 to 10, where 0 means the response is completely unreasonable, and 10 means the response is perfect. Please also provide a brief explanation of your score.\newline

Give your score in the following format:\newline
<score>\{\{Your score\}\}</score>\newline

Question: \{question\}\newline
Response: \{response\}
\end{promptbox}
\caption{GPT Evaluation template}
\label{fig:evaluate_prompt}

\end{figure*}

\section{Dataset Details}
\label{sec:dataset_details}
In this section, we elucidate the seed dataset and the evaluation sets.
We also clearly denote the involved dataset's usage during training and evaluation and their corresponding category in Table~\ref{tab:dataset_usage}.
We divide the used 16 training datasets into the following five dimensions:
\begin{enumerate}
    \item \textbf{Long Context QA}: This dimension enables \mone to capture useful information from the given context and response with long-chain reasoning. This dimension covers BioMRC~\citep{pappas-etal-2020-biomrc}, HeadQA Topic Classification~\citep{vilares-gomez-rodriguez-2019-head,wu2024towards}, and HealthFact~\citep{kotonya-toni-2020-explainable-automated}
    \item \textbf{Knowledge-Intensive QA}: This dimension teaches \mone to use long-chain reasoning to answer knowledge-intensive problems, which covers MedQA~\citep{jin2021disease}, MedMCQA~\citep{pal2022medmcqa}, and PubMedQA~\citep{jin-etal-2019-pubmedqa}.
    \item \textbf{Bio-Medical QA}: This part leverages general data in bio-medicine domains to enhance the generality of \mone, which includes SciQ~\citep{welbl-etal-2017-crowdsourcing}, Evidence Inference~\citep{deyoung-etal-2020-evidence} and Head QA~\citep{vilares-gomez-rodriguez-2019-head}.
    \item \textbf{Medical Natural Language Inference}: This dimension prompts \mone to discriminate biomedical research concepts and corresponding descriptions, which contain PubHealth~\citep{kotonya-toni-2020-explainable-automated}, Medical Question Pair (MQP; \citet{mccreery2020effective}), and catalonia-independence-corpus~(CIC; \citet{zotova-etal-2020-multilingual}).
    \item \textbf{Diagnosis QA}: This dimension is related to real-world clinical scenarios, including disease diagnosis and classification and drug related questions. We choose Covid-19 Classification~\citep{covid19classification2020}, Drug-Dose Extraction, Adverse Drug Event Classification~\citep{huynh-etal-2016-adverse,wu2024towards} and DDX-Plus~\citep{Fansi_Tchango2022}.. 
\end{enumerate}

The descriptions of each training and evaluation dataset are presented below:
\begin{enumerate}
    \item MedQA~\citep{jin2021disease} is a widely used benchmark for evaluating AI systems in medical question answering, featuring multiple-choice questions from professional medical licensing exams such as the USMLE and exams from China and Taiwan. We adopt its 5-options English version, taking its training set as seed data and 1,273 test problems as the evaluation benchmark.
    \item PubmedQA~\citep{jin-etal-2019-pubmedqa} is a specialized benchmark for biomedical question answering, consisting of question-answer pairs derived from PubMed abstracts. It focuses on yes/no/maybe questions that require reasoning over biomedical literature. We use the human-labeled question set and split the training set and test set, with both 500 problems for evolution and evaluation, respectively. Note that we do not include relevant contexts before questions, challenging models' internal knowledge comprehension.
    \item MedMCQA~\citep{pal2022medmcqa} is a large-scale benchmark for medical question answering, featuring over 194,000 multiple-choice questions sourced from Indian medical entrance exams and other educational resources. It spans a wide range of medical topics, including anatomy, pharmacology, and pathology, and is designed to evaluate the reasoning and knowledge application skills of AI systems in a clinical context. The test set contains 4,183 problems.
    \item MMLU~\citep{hendrycks2021measuring} is to measure LLM's multitask accuracy, which contains 14,421 problems. The test covers 57 tasks including elementary mathematics, US history, computer science, law, and more. We select its medical-related problems, resulting in a test set with 1,561 problems.
    \item BioMRC~\citep{pappas-etal-2020-biomrc} is a collection of medical-related question-answer pairs, specifically designed for the evaluation of machine reading comprehension (MRC) tasks in the biomedical domain. It is derived from a wide range of medical texts, including clinical notes, research papers, and medical textbooks. The dataset contains a series of questions and corresponding answers, where the answers are extracted from relevant passages of text. We use its 6,250 test set as the evaluation set.
    \item HeadQA~\citep{vilares-gomez-rodriguez-2019-head} is a specialized medical question-answering dataset designed to evaluate models in the context of neurology and head-related disorders. It consists of a collection of questions paired with answers derived from a variety of clinical notes, medical reports, and other head-related health data sources. 
    \item DDX-Plus~\citep{Fansi_Tchango2022} is a comprehensive medical diagnostic dataset designed to assist in the development and evaluation of machine learning models for differential diagnosis in clinical settings. \blue{It consists of clinical cases, where each case includes a set of symptoms, patient history, physical examination findings, and diagnostic questions, along with a list of potential diagnoses ranked by their likelihood. The diverse set of cases in the dataset spans multiple medical specialties, making it an ideal resource for creating models capable of assisting healthcare professionals in making informed diagnostic decisions.} Due to its huge test set (over 100,000 test instances), we randomly select 2,000 items for evaluation.
    \item SciQ~\citep{welbl-etal-2017-crowdsourcing} is a scientific question-answering dataset designed to assess machine learning models in answering factual questions across a wide range of scientific domains. It consists of over 13,000 questions derived from scientific literature, including topics in physics, biology, chemistry, and earth sciences, among others. Each question is paired with a correct answer and is supported by a passage of text from which the answer is extracted.
    \item Evidence Inference~\citep{deyoung-etal-2020-evidence} is a collection designed to evaluate machine learning models on their ability to infer logical conclusions from evidence presented in the form of textual information. This dataset consists of structured pairs of premises (evidence) and hypotheses, where the goal is for models to determine the logical relationship between them—whether the hypothesis is supported, contradicted, or is neutral with respect to the provided evidence. Typically used for tasks such as textual entailment or natural language inference (NLI), the dataset includes a variety of complex scenarios across multiple domains, including law, healthcare, and science, where reasoning based on available evidence is crucial.
    \item PubHealth~\citep{kotonya-toni-2020-explainable-automated} is a comprehensive dataset for explainable automated fact-checking of public health claims. Each instance in the PUBHEALTH dataset has an associated veracity label (true, false, unproven, mixture). Furthermore, each instance in the dataset has an explanation text field. The explanation is a justification for which the claim has been assigned a particular veracity label. We construct two different test sets. Healthfact is to directly predict whether a given instance is true/false/unproven/mixture. The other, Pubhealth, is to predict whether the instance sentence and the given explanation express the same meaning.
    \item Medical Question Pair~\citep{mccreery2020effective} contains a dataset of 3,048 similar and dissimilar medical question pairs hand-generated and labeled by Curai's doctors. Models should clarify whether the two given questions are similar or not.
    \item Catalonia-independence-Corpus~\citep{zotova-etal-2020-multilingual} is a dataset built for stance detection in Twitter for the Catalan and Spanish languages, with the aim of facilitating research on stance detection in multilingual and cross-lingual settings.
    \item Covid-19 Classification~\citep{covid19classification2020} is an extension of the Hedwig library and contains all necessary code to reproduce the results of some document classification models on a COVID-19 dataset created from the LitCovid collection.
    \item Adverse Drug Event~\citep{huynh-etal-2016-adverse} is critical for developing automated systems that can support clinicians in identifying harmful drug reactions, potentially reducing healthcare costs, and enhancing patient safety. Given the increasing volume of clinical data, this dataset plays a key role in advancing AI-driven drug safety research and improving the overall quality of healthcare. We build Drugdose extraction test set to benchmark models to extract the exact dose of a specific drug.
    \item \blue{SEER~\citep{dubey2023using} is purposed for treatment planning because it contains key clinical variables that directly inform therapy decisions (e.g., tumor size, nodal status, hormone receptor status). LLMs must answer the most appropriate suggestion from the following list ['Intraoperative rad with other rad before/after surgery', 'Intraoperative radiation', 'No radiation and/or cancer-directed surgery', 'Radiation after surgery', 'Radiation before and after surgery', 'Radiation prior to surgery', 'Surgery both before and after radiation'] based on patient summarization, simulating real-world tumor board decisions. }
\end{enumerate}

\begin{table}[tbp]
  \centering
   \resizebox{0.46\textwidth}{!}{%
    \begin{tabular}{clcc}
    \toprule
    \multicolumn{1}{l}{\textbf{Category}} & \textbf{Dataset} & \textbf{Train} & \textbf{Test} \\
    \midrule
    \multirow{4}[2]{*}{Diagnosis QA} & ADE   & Yes   & No \\
          & Covid-19 CLS & Yes   & No \\
          & DrugDose & Yes   & Yes \\
          & DDXPlus & Yes   & Yes \\
          & SEER  & No    & Yes \\
    \midrule
    \multirow{3}[2]{*}{Medical NLI} & PubHealth & Yes   & Yes \\
          & CIC   & Yes   & No \\
          & MQP   & Yes   & No \\
    \midrule
    \multirow{3}[2]{*}{Long Context QA} & BioMRC & Yes   & Yes \\
          & HealthFact & Yes   & Yes \\
          & HeadQA Topic CLS & Yes   & No \\
    \midrule
    \multirow{3}[2]{*}{BioMedical QA} & HeadQA & Yes   & No \\
          & Evidence Extraction & Yes   & No \\
          & SciQ  & Yes   & No \\
    \midrule
    \multirow{5}[2]{*}{Knowledge QA} & MedQA & Yes   & Yes \\
          & MedMCQA & Yes   & Yes \\
          & PubMedQA & Yes   & Yes \\
          & MMLU  & No    & Yes \\
          & BioASQ & No    & Yes \\
    \bottomrule
    \end{tabular}%
    }
      \caption{Medical datasets usage during training and evaluation. ``CLS'' denotes classification.}
  \label{tab:dataset_usage}%
\end{table}%

\section{Hyperparameters}
\label{sec:hyperparameters}
\subsection{Data Synthesis}
For each node expansion, we simultaneously generate 3 different responses with the same prompt.
We set the generation temperature to 1.
The stop tokens are set to $\{\texttt{Step k:}\mid k=1,2,\cdots 100\}$ to ensure that each node represents a single reasoning step.
We use the first sample in MedQA as the one-shot example and prompt GPT-4o to generate step-by-step outputs.

\subsection{Self-Training of Policy and PRM}
\label{sec:prm_tuning_details}
We use 8xNVIDIA A100 GPUs and the overall training consumes 14h.
\paragraph{Policy tuning}
We use trl\footnote{https://huggingface.co/docs/trl/index} as the training framework.
We first use vanilla \texttt{SFTTrainer} to train the policy model.
We set the warmup ratio to 0.03 and the max sequence length to 8192.
The batch size is set to 128 and the learning rate is set to 1e-6.
After that, we use \texttt{DPOTrainer} to further fine-tune the policy model.
We set the learning rate to 5e-8 and the batch size to 128.


\paragraph{PRM tuning}
We use \texttt{PRMTrainer} of trl to train the PRM model.
We use LoRA to fine-tune the PRM, where the lora rank is set to 32 and lora alpha set to 64.
The learning rate is set to 5e-5.
\blue{
For a single step $s_k$, the input for PRM is the concatenation of all steps up to the current step, namely:}
\begin{align}
    P&=s_0\oplus s_1\oplus\cdots\oplus s_k \\
    \hat{y}&=V_{\theta}(P; x) 
\end{align}
\blue{
This input models a step's value with causal relationships between steps, preventing local optima learning.
}

\subsection{Evaluation}
For evaluation, the temperature is set to $1.0$ and top\_p is set to $0.9$ for multiple sampling settings.
For the comparison between policy models, the temperature is set to 0 and the greedy decoding method is adopted to avoid variable results.
The max generation tokens are set to 8,192.
Among the three presented decoding mechanisms, CoT~\citep{wei2022chain} directly prompts models to generate a long reasoning chain and outputs the answer with ``The answer is \{answer\}'' for the convenience of answer extraction.
Self-Consistency~\citep{wang2023selfconsistency} generates $N=32$ samples for a given problem, and we select the one whose answer appears most times among the $N$ outputs.
We use exact match~(EM) to measure the performance.
Specifically, we extract the contents following the last ``The answer is'' template to match the self-reflection thinking style, and perform appropriate post-processing to derive a final prediction.
No matter what the contents the model has generated, we manually append ``The answer is'' to the end of the generation, and prompt the model to continue generation, with a small token limit (20) and preset logit bias to ensure the generation falls into candidate tokens.
For multiple-choice problems, we directly choose the first character of prediction phrases and measure whether the ground truth is equal to the prediction.
For close-ended generation tasks, we remove quotes and turn the prediction and the ground truth into lowercase phrases.
After that, we check whether the ground truth phrases exist in the prediction phrases.

\subsection{Training Details of Distillation and RL}
In this section, we elucidate the implementation details of distillation and RL.
\paragraph{Distillation} For Distillation method, we fine-tune Llama3.1-8B with 2K training data\footnote{\url{https://huggingface.co/datasets/SPIRAL-MED/o1-journey-Ophiuchus}} released by~\citet{huang2025o1}, which combined with the questions in MedQA and corresponding response generated by o1~\cite{jaech2024openai}. We adopt LoRA~\citep{hu2022lora} and set the rank $r$ to 16 and alpha $\alpha$ to 32 for fair comparisons. For fine-tuning parameters, we set the learning rate to 2$e$-6 and batch size to 128.

\paragraph{RL}  
We follow \citet{guo2025deepseek} to use Group Relative Policy Optimization (GRPO; \citet{shao2024deepseekmath}) to conduct RL training. 
We set the number of generations to 10 and the learning rate to $1e-6$.
We adopt ZeRO-3~\citep{rajbhandari2020zero} to save memory and conduct full fine-tuning in one 8xA100 machine.
The batch size is set to 4 per GPU.
For the adopted prompt, we use the same prompt illustrated in DeepSeek-R1-zero, and use \texttt{<think></think><answer></answer>} to learn the slow-thinking output style.
We use accuracy reward and format reward, where
\begin{align}
    r=\begin{cases}
        1 & \text{Correct answer with correct format} \\
        0 & \text{Incorrect answer with correct format} \\
        -1 & \text{Incorrect format} \\
    \end{cases}
\end{align}


\section{Best-of-N Details}
\label{sec:best_of_n_details}
In this section, we elucidate the fast inference using Best-of-N (BoN) evaluation with the PRM.
Specifically, the policy model generates $N$ responses $\{y_i\mid i\in[1,N]\}$ simultaneously using the inference engine (vLLM; \citet{kwon2023efficient}).
After that we split each response $y_i$ into $K$ steps: $[s_i^1,s_i^2,\cdots,s_i^k]$ where $k\in[1,K]$ using predefined step separator \texttt{Step}.
We form the step input to the PRM $(x,S_i^k)$ as $S_i^k= s_i^1\oplus s_i^2\oplus\cdots\oplus s_i^k$ and obtain each step's value $v_i^k=\mathrm{softmax}(V_{\theta}(x; S_i^k))$ using a forward pass. 
The value is the probability of predicting the step to be correct, which is in the range $[0,1]$.
Finally, the whole response value $v_i=\min_{k=1}^K v_i^k$ is taken from the minimum value of each step and we select the response with the highest value.


\section{Case Study}

\begin{figure*}[tbp]
    \centering
    \includegraphics[width=0.8\linewidth]{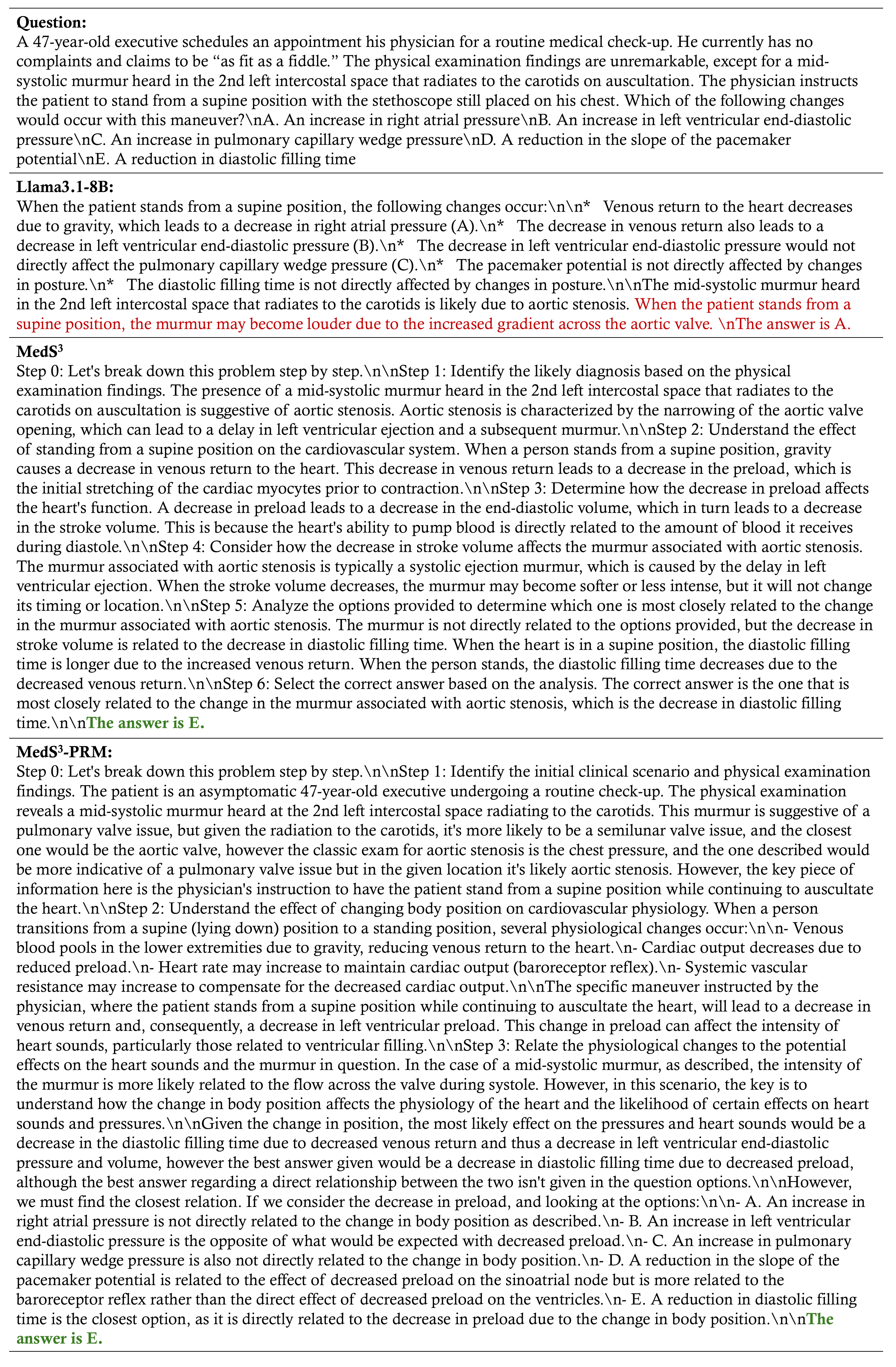}
    \caption{Cases of MedQA}
    \label{fig: medqa_case}
\end{figure*}

\begin{figure*}[tbp]
    \centering
    \includegraphics[width=0.9\linewidth]{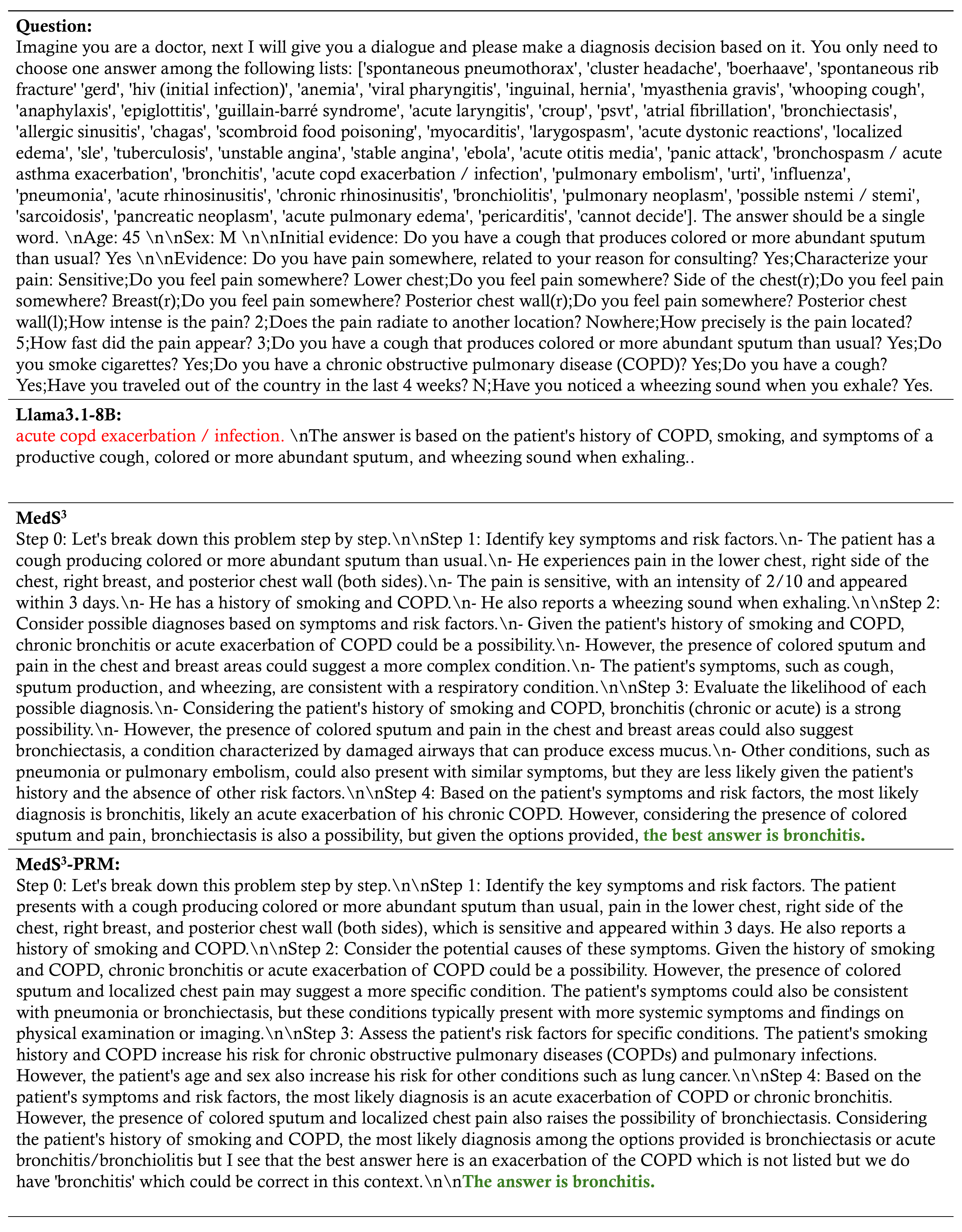}
    \caption{Cases of DDXPlus}
    \label{fig: ddxplus_case}
\end{figure*}

\begin{figure*}[tbp]
    \centering
    \includegraphics[width=0.9\linewidth]{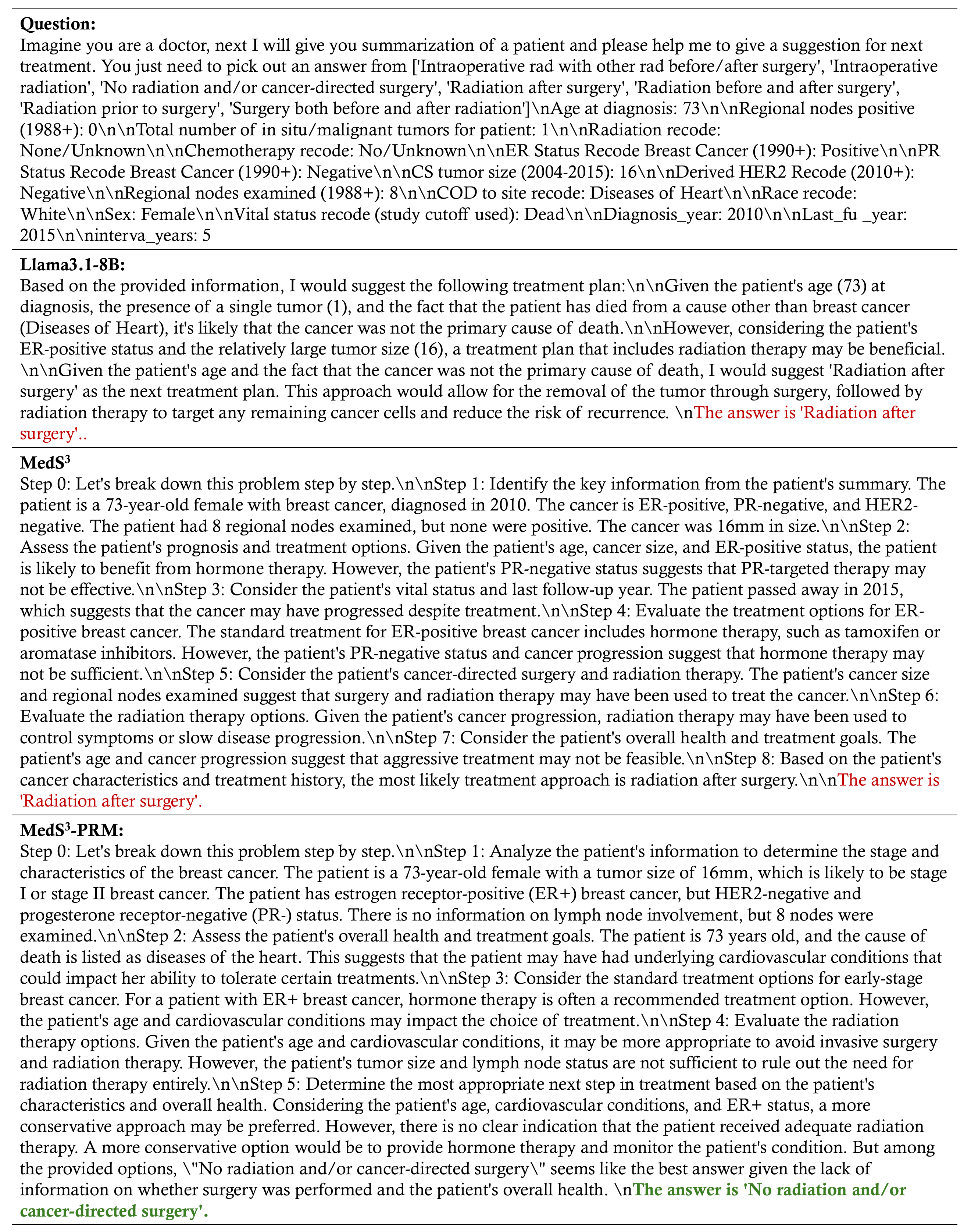}
    \caption{Cases of SEER}
    \label{fig: seer_case}
\end{figure*}

\end{document}